\documentclass[10pt,twocolumn,letterpaper]{article}

\usepackage{cvpr}
\usepackage{times}
\usepackage{epsfig}
\usepackage{graphicx}
\usepackage{amsmath}
\usepackage{amssymb}

\usepackage[export]{adjustbox}
\usepackage{booktabs}
\usepackage[all]{nowidow}
\usepackage{float}
\usepackage{multirow}
\usepackage{enumitem}
\usepackage{stfloats}
\usepackage{comment}
\usepackage[super]{nth}
\usepackage[caption=false]{subfig}
\usepackage{arydshln}

\newcommand\mypara[1]{\vspace{2mm}\noindent\textbf{#1}}

\def\eg{\emph{e.g}\onedot}
\def\Eg{\emph{E.g}\onedot}

\def\ie{\emph{i.e}\onedot}
\def\Ie{\emph{I.e}\onedot}

\newcommand{\roiOne}{\textsc{ROI1}\xspace}
\newcommand{\roiTwo}{\textsc{ROI2}\xspace}
\newcommand{\roiThree}{\textsc{ROI3}\xspace}

\usepackage[pagebackref=false,breaklinks=true,colorlinks,bookmarks=false]{hyperref}
\usepackage{multibib}
\newcites{Supp}{References} %

\cvprfinalcopy %

\def\cvprPaperID{3} %

\ifcvprfinal\fi
\begin{document}

\title{ResDepth: Learned Residual Stereo Reconstruction}

\author{Corinne Stucker\qquad Konrad Schindler \vspace{1mm}\\
Photogrammetry and Remote Sensing, ETH Zurich, Switzerland\\
{\tt\small \{firstname.lastname@geod.baug.ethz.ch\}}
}

\maketitle
\thispagestyle{empty}

\begin{abstract}
We propose an embarrassingly simple but very effective scheme for high-quality dense stereo reconstruction: (i)~generate an approximate reconstruction with your favourite stereo matcher; (ii)~rewarp the input images with that approximate model; (iii)~with the initial reconstruction and the warped images as input, train a deep network to enhance the reconstruction by regressing a residual correction; and (iv)~if desired, iterate the refinement with the new, improved reconstruction. The strategy to only learn the residual greatly simplifies the learning problem. A standard Unet without bells and whistles is enough to reconstruct even small surface details, like dormers and roof substructures in satellite images. We also investigate residual reconstruction with less information and find that even a single image is enough to greatly improve an approximate reconstruction. Our full model reduces the mean absolute error of state-of-the-art stereo reconstruction systems by $>$50\%, both in our target domain of satellite stereo and on stereo pairs from the ETH3D benchmark.
\end{abstract}
\section{Introduction}
Dense stereo reconstruction is a classical task of computer vision with a rich history and an elementary building block of 3D perception. The problem statement is simple: given two images with overlapping fields of view and known relative pose, find a 3D scene that is photo-consistent with both views. Efficient solutions exist and form the basis for a wide range of operational systems, ranging from large-scale topographic reconstruction to industrial machine vision and mobile robotics.

\begin{figure}[t!]
\centering
\includegraphics[trim=100 100 130 80,clip,width=\columnwidth]{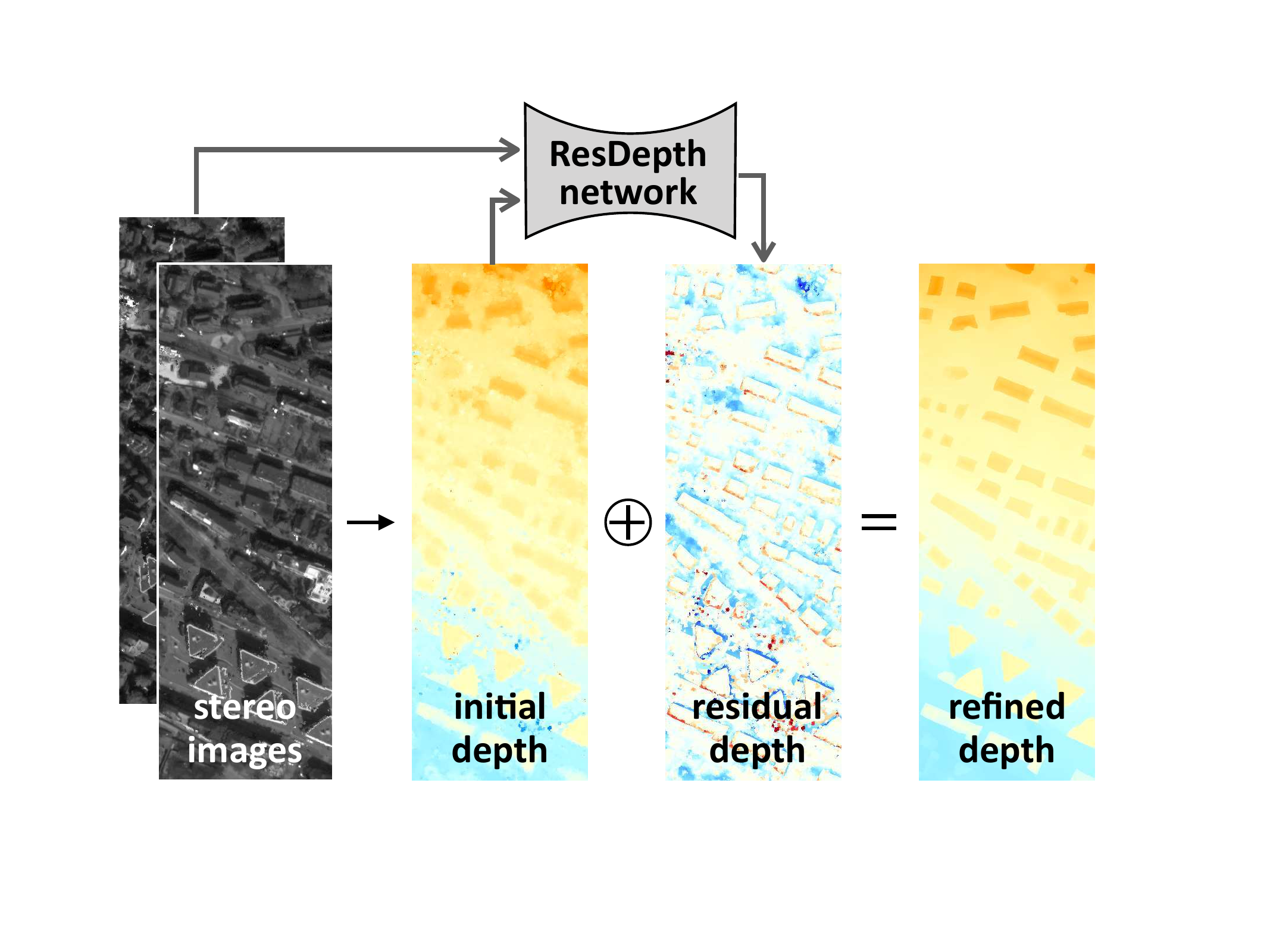}
\caption{Instead of learning stereo reconstruction from scratch, our \emph{ResDepth} network is trained to refine an initial depth (respectively, height) map with the help of stereo images. Our main message is that \emph{stereo guidance} greatly improves the refined depth.}
\label{fig:teaser}
\end{figure}

For many practical applications, maximising photo-consistency across all pixels is not enough to solve dense stereo. Rather, one must also impose a suitable prior on the 3D scene. Classical stereo algorithms~\cite{kolmogorov2001computing,hirschmuller2005accurate} typically include an explicit preference for piece-wise smooth surfaces. Since that prior is rather unspecific and knows very little about the structure of the observed scene, the resulting surface models must in practice still be cleaned up. For example, in topographic mapping, it is common to enhance building shapes with heuristic rules; in mobile robotics, reflecting glass must be detected and cleaned up in post-processing; \etc. 

The need to capture complex, soft prior expectations about the world, which are hard to formulate explicitly, naturally calls for a machine learning approach. Indeed, several authors have recently proposed to learn stereo matching, \ie, to design a deep encoder-decoder network that maps the input images to a depth map~\cite{mayer2016large,kendall2017end,chang2018pyramid,treible2018learning}. We argue that, while conceptually elegant, such a purely learning-based solution may be inefficient because it has to learn many things from data that are already well captured by existing stereo methods. Importantly, classical stereo matching algorithms are very robust in the sense that their outputs are usually correct as a coarse, \emph{global} estimate of the scene surface but may suffer from \emph{local} biases and errors. Their main shortcoming is a lack of prior knowledge about the observed world beyond simplistic (piece-wise) smoothness.

In this paper, we thus advocate a \emph{residual learning} strategy. We reconstruct an approximate surface with a standard stereo matcher---we use a conventional one, but it could be a learned one too---that may have certain biases but robustly produces passable stereo models. We then train a deep network to upgrade that initial estimate by regressing an additive residual correction, see Fig.~\ref{fig:teaser}. The input to the proposed \textit{ResDepth} network are two input images warped onto the initial depth (respectively, height) map and the initial depth map itself. In this way, the network has access to not only the images but also to the initial surface model and can concentrate on the part of the problem for which it is most needed: its task becomes to intervene where the assumptions built into the initial stereo algorithm fail and a more intricate prior is needed that must be learned from data. Intuitively, our network leverages the correlation between intensity patterns in the warped synthetic images and depth patterns in the initial reconstruction. Wherever the warped images are \textit{not} photo-consistent or depth and texture discontinuities do not coincide, the network refines the initial depth by applying an additive residual correction. Regressing residual corrections that are mostly small is much easier than learning the entire depth estimation process---this is just another of the ubiquitous optimisation strategies to start from an approximate solution and refine it. Further, it is also easier to choose the right prior locally for a specific region whose 3D geometry is already roughly known. For instance, a house-sized protrusion in a city model is a strong indication that the surface might conform to a common roof shape. Finally, the proposed approach makes it easy to tune the reconstruction to the user's needs. \Eg, one can train \emph{ResDepth} to remove trees from the initial surface model simply by supervising it with a city model without trees. Such a filter would be a lot more difficult to construct without an initial reconstruction to guide both tree detection and "inpainting" of the correct ground height. An alternative interpretation of our method is as a learned enhancement filter for depth maps guided by the original image content.

In summary, our \textbf{contributions} are: \emph{(i)} We propose \textit{ResDepth}---a residual network that leverages stereo guidance to improve an initial depth obtained from conventional stereo matchers. ResDepth is effective, efficient, and applicable to large-scale scenes. \emph{(ii)} In our target domain of satellite stereo reconstruction, we show significant improvements over state-of-the-art, including a 2.5$\times$ reduction of the median error, more accurate building shapes and outlines, and faithful reconstruction of surface details like dormers.

\section{Related Work}

\mypara{Conventional Stereo Matching}
Traditional stereo methods find a dense set of correspondences that have high photo-consistency, while at the same time forming a (piece-wise) smooth surface. The key issue is to efficiently approximate the smoothness prior using for instance graph cuts~\cite{kolmogorov2001computing}, dynamic programming~\cite{hirschmuller2005accurate} or the PatchMatch method~\cite{bleyer2011patchmatch}. 
To handle high-resolution images, \eg, in aerial mapping, these methods are often employed iteratively in a spatial pyramid scheme~\cite{rothermel2012sure}, whose later iterations can be seen as a refinement of a coarser initial solution. Other stereo algorithms are by design iterative, including many variational schemes~\cite{slesareva2005optic,ben2007variational} and methods based on mesh surfaces~\cite{lengagne2000_3d}.

\mypara{Deep Stereo Matching}
In the last few years, the focus has been on stereo methods that harness the power of deep learning. While early attempts only learned to measure patch similarity within a conventional optimisation~\cite{zbontar2016stereo}, more recent methods use encoder-decoder architectures to directly output disparity maps~\cite{mayer2016large,kendall2017end,chang2018pyramid}. Some of the latest methods can also handle high-resolution images~\cite{tulyakov2018practical,yang2019hierarchical}. Most closely related to our work are recent, rather \mbox{complex} stereo architectures like~\cite{pang2017cascade,liang2018learning,jie2018left} that internally split the computation into a first, coarse disparity estimation and a subsequent refinement. However, these methods do not investigate the influence of the refinement step in isolation and seem to imply that end-to-end, deep integration of the cascade is crucial. We show that this is \emph{not} strictly necessary and concentrate on a detailed analysis of the refinement part. The critical step for high-quality reconstructions appears to be the learned refinement, which can be accomplished with a simple standard architecture that needs much less training data. As initialisation, standard stereo methods are sufficient. In our target domain of satellite imaging, conventional matchers still dominate deep stereo networks, \eg, in the \textit{IARPA Multi-View Stereo 3D Mapping Challenge 2016}~\cite{iarpa2016challenge} all top-performers are variants of semi-global matching~\cite{hirschmuller2005accurate}, ahead of deep methods like~\cite{treible2018learning}.

\mypara{Filtering and Refinement of Depth Maps}
Several works have looked at ways to improve an initial 3D surface. \mbox{\cite{vu2011high,blaha2017semantically}} iteratively deform a surface mesh to maximise photo-consistency. \cite{gidaris2017detect} exploit monocular image information to detect and refine regions of incorrect disparity, whereas \cite{jie2018left} use a left-right comparison to guide the refinement.
Unlike~\cite{gidaris2017detect,jie2018left}, we prefer \emph{not} to select regions to be refined explicitly, which creates a point of failure. Instead, we update everywhere, with update 0 where not needed. In the context of topographic mapping, it is common to refine buildings by fitting parametric models~\cite{haala1997integrated,lafarge2011building}. 

\mypara{Guided Depth and Disparity Enhancement} We see previous works such as~\cite{pang2017cascade,batsos2018recresnet} as most closely related to ours. While these methods adopt a coarse-to-fine scheme for the refinement, we leverage a large receptive field and exploit local and global long-range context to estimate a one-shot residual update at the resolution of the input. \cite{batsos2018recresnet} propose a recurrent residual network guided by a \emph{single} image to refine initial disparities and claim that a binocular setup would not be helpful. In contrast, we build a two-view system and experimentally show that \emph{stereo} guidance does improve performance. \cite{pang2017cascade} use binocular image guidance and the difference between the left and warped right view as additional input to the refinement network. We refrain from using the latter input as this information is not meaningful for images captured under strong lighting differences---like satellite images from different days. Furthermore, we do not use an explicit correlation layer. Recently, \cite{bittner2018dsm} propose to learn a general prior for enhancing digital elevation models, including a version guided by a \emph{single} ortho-image~\cite{bittner2019dsm,bittner2019late}.
\section{Method}

\begin{figure*}[ht]
\centering
\includegraphics[width=0.975\textwidth]{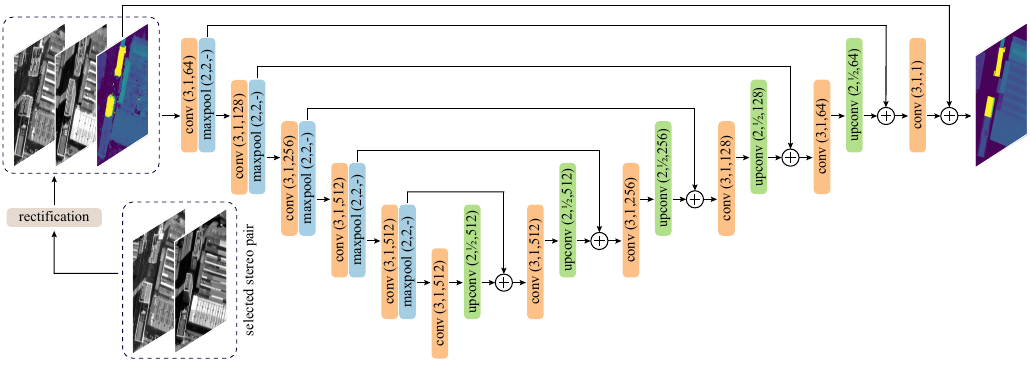}
\caption{ResDepth-\textit{stereo} network architecture for satellite images. Each convolutional layer except the last one is followed by batch normalization and ReLU activation function. The numbers in parentheses indicate kernel size, stride, and the number of kernels. Note the long residual connection that directly adds the input depth to the output of the last decoder layer to force the network to learn residual depths instead of absolute depths. For close-range data, we add two down- and upsampling levels to account for the larger input patch size.}
\label{fig:network}
\end{figure*}

Our goal is dense surface reconstruction by stereo matching. Rather than designing a deep encoder-decoder network that learns to map the input images to a depth map in an end-to-end manner, we start with a coarse reconstruction and train a deep network to upgrade that initial estimate by regressing an additive residual correction. Our residual depth regression network \emph{ResDepth} is based on a simple Unet~\cite{ronneberger2015u} and uses stereo information as guidance, see~Fig.~\ref{fig:network}.

Our method starts from images with overlapping fields of view and known camera poses. For simplicity, the following explanations assume a binocular stereo setup. Note that extending ResDepth to a fixed number of views $>$2 is straightforward by simply adding more input channels to the network. We describe our technical approach from a general point of view and highlight conceptual details tailored to our target domain of satellite stereo.

\subsection{Initial Reconstruction}
To play to the strength of convolutional networks, the coarse initial reconstruction should be parametrised over a regular 2D grid. In general, the reconstruction is represented as a depth map in the camera coordinate system of one of the views. In the special case of aerial or satellite imaging, a more natural choice is a \emph{digital elevation model} (DEM), \ie, a raster of height values along the gravity axis.

\subsection{Image Rectification}
For further processing in ResDepth, the images should be aligned with the initial depth map. The first view used to parametrise the depth map is aligned by design. The pixel-wise alignment of the second view is achieved by rewarping the image to the camera coordinate system of the first view using the initial disparity. For satellite images, this corresponds to independently ortho-rectifying the images.

Note that rewarping compensates the influence of the viewing geometry to an even greater extent than epipolar rectification. The disparities in the warped images are, by construction, small everywhere except at large depth errors. Thus, no model capacity is wasted on learning stereo reconstruction across a wide range of permissible disparities.

If the reconstruction were perfect, the two input views would be perfectly aligned after rewarping, and hence maximally photo-consistent. Therefore, discrepancies between the two warped images provide a signal of how to improve the depth. The only regions where a correct depth map would not achieve photo-consistency are those that are occluded in one view. We deliberately do not perform ray-casting to account for occlusions during the rewarping process. Instead, we render duplicate textures if the corresponding rays intersect the surface twice, leading to photometrically inconsistent, systematically displaced copies of nearby textures. The rationale is that these very systematic patterns provide even stronger evidence about the surface shape than empty pixels. This effect is particularly visible for tall buildings in satellite images (see ~Fig.~\ref{fig:network}).

\subsection{Network Architecture}
Fig.~\ref{fig:network} depicts the network architecture of ResDepth. We found that a fairly standard Unet~\cite{ronneberger2015u} works well. We use 5~levels for satellite images (input patch size 128$\times$128) and 7~levels for close-range images (input patch size 512$\times$512), such that in both cases the bottleneck has dimensions 512$\times$4$\times$4. Each encoder level consists of the sequence \emph{3$\times$3~conv -- batch norm -- ReLU -- 2$\times$2 max pool}. The decoder levels are similar, except that max-pooling is replaced by up-convolution with stride $\frac{1}{2}$. All inputs are stacked into a single multi-channel image and fed to the network.

The main motivation for our residual stereo method is to reconstruct crisp crease edges and depth discontinuities and to align them with the image content. In that context, an important feature of a Unet-type architecture is the exhaustive set of skip connections from encoder to decoder levels of the same resolution, which make sure no high-frequency detail is lost. Recall that in our case we add a long residual connection that directly adds the input depth to the output of the last Unet layer, so that the network only learns what must be added to the input depth to get to the ground truth.

\subsection{Network Variants}
Our method can be regarded as an image-guided depth enhancement filter. That view leads to the question of how much each input channel contributes to the refined depth, and whether all inputs are indeed needed to achieve the desired effect. Moreover, there may be situations where not all inputs are available, \eg, one may be faced with the task of improving an existing DEM for which one has access to a single image only but not to the stereo coverage.

Therefore, we construct several ResDepth variants that differ in number and combination of input modalities but are otherwise identical. In particular, we keep the network architecture fixed and train each variant using the same training settings and data. We then compare the performance to systematically investigate how the input modalities affect the quality of the refined depth. The network configuration based on stereo guidance represents our full model, which we refer to as ResDepth-\textit{stereo}.

ResDepth-\textit{mono} does not use the second input image, similar to~\cite{batsos2018recresnet,bittner2019dsm, bittner2019late}. Therefore, it has no access to stereo disparities but can still use the monocular image information to enhance the depth map. Potentially, this includes both low-level and high-level information. Low-level image edges may serve to sharpen and localise depth edges and jumps, in the spirit of the guided filter~\cite{he2010guided}. High-level information implicit in the image, like semantics and layout, can also be valuable, \eg, it may serve to distinguish large trees from small buildings.

ResDepth-\textit{0} does not use any image information at all. It merely learns a prior on the structure of depth images and corrects unlikely configurations of depth values without conditioning on image evidence, as in~\cite{bittner2018dsm}. In the above analogy, it can be thought of as a clever, learned bilateral filter~\cite{tomasi1998guided} with the additional capability to recognise and exploit long-range correlations such as straight gable lines.

As a sanity check for the claim that residual depth is easier to learn than full stereo matching, we formulate a variant without the initial depth map as input. \Ie, the Unet is fed the two warped images only and trained to output the full depth map. For simplicity, note that we use the same warping as for ResDepth variants that include the initial depth map. Thus, the task is, in fact, easier than stereo matching from scratch. We refer to this variant as Unet-\emph{stereo}.

In the opposite direction, an interesting extension is to iterate the residual correction. ResDepth-\textit{stereo\textsubscript{iter}} uses the output depth of ResDepth-\textit{stereo} as input, warps the input images with the new, improved depth map, and trains another ResDepth-\textit{stereo} network to further reduce the new, smaller depth errors. We note that, in principle, it is possible to concatenate the two ResDepth-\textit{stereo} networks and train them end-to-end since the image warping is differentiable. While shared weights across unrolled network iterations~\cite{Hur2019IRR} are certainly more elegant, it is unlikely that doing so will in practice lead to better results.

Further, we design a more sophisticated ResDepth-\textit{stereo} variant that generalises across viewing directions and lighting conditions between views. This property is particularly interesting for the task of updating an existing DEM with a stereo pair that might not comply with the viewing or lighting conditions of the images used during training. We pragmatically realise the generalised ResDepth-\textit{stereo} variant by independently combining the same DEM patch with different ortho-rectified stereo pairs during training.

\section{Experimental Evaluation}

\subsection{Satellite Images}
Our main application is large-scale urban 3D modelling from high-resolution satellite images. Stereo methods applied to satellite imagery produce digital elevation models (DEM) with sub-meter spatial resolution~\cite{ozcanli2015comparison,facciolo2017automatic,leotta2019urban}. Despite the impressive resolution, reconstructions based on satellite images are comparatively noisy, blurred, and often incomplete due to restrictions on the viewing directions. This, in turn, makes downstream modeling difficult. We see ResDepth as a simple yet effective way to improve satellite stereo reconstructions by combining stereo correspondence with learned priors for building shapes and layout, treatment of vegetation, systematic shadow patterns, \etc

\begin{figure}[t]
\centering
\includegraphics[scale=0.7,trim={0 3 0 6},clip]{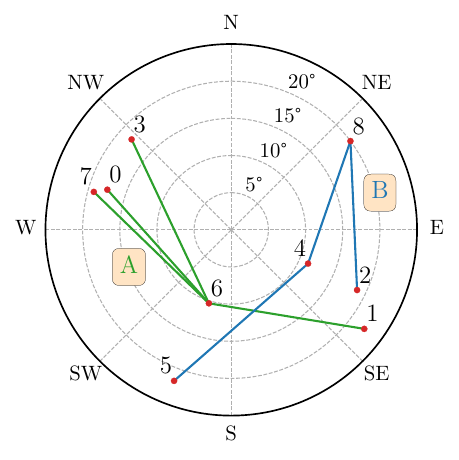}
\caption{Satellite position (red marker) for each image acquisition over Zurich, Switzerland. The circular orientation refers to the azimuth angle and the radial direction to the off-nadir angle. The acquisitions are sorted and labelled in ascending order (from older to newer acquisitions). We show the partitioning of the stereo pairs that conform to our selection criteria (see text for details) into two mutually exclusive groups $A$ (green) and $B$ (blue). Stereo pairs are connected by a line.}
\label{fig:stereopairs} 
\end{figure}

We use WV-2 and WV-3 images acquired over Zurich, Switzerland (see Fig.~\ref{fig:stereopairs}). The area covered is 1.5$\times$1.5$\,$km$^\text{2}$ and includes industrial and residential districts. The images were captured between 2014 and 2018, with 1.5~months as the shortest time interval between acquisitions.

\mypara{Initial Reconstruction}
As usual in satellite imaging, only the panchromatic channel is used for reconstruction since it is recorded at a higher resolution (ground sampling distance of 0.46$\,$m at nadir). We use a re-implementation of state-of-the-art hierarchical semi-global matching~\cite{rothermel2012sure} tailored to satellite images to generate the initial DEM. Its grid spacing is 0.25~m, leading to a total of $\approx 3.7\cdot 10^7$ pixels.

\mypara{Image Selection}
Based on~\cite{facciolo2017automatic}, we use a simple heuristic to define the stereo pair used as input to ResDepth. Starting from all possible pairs, we first eliminate all pairs whose intersection angle is smaller than 10 or larger than 28~degrees or that contain an image with an incidence angle beyond 40~degrees. From the remaining stereo pairs, we choose the one with the smallest time difference (modulo 180 days).

To test generalisation, we split the selected stereo pairs into two mutually exclusive groups $A$ and $B$ such that each group contains along-track and across-track stereo pairs, see Fig.~\ref{fig:stereopairs}. We use the stereo pairs in $A$ for training and the ones in $B$ for testing (and vice versa).

\mypara{Ground Truth}
We use the publicly available 2.5D CAD city model of Zurich~\cite{zurich3Dmodel} to render the ground truth DEM of the same area. The city model was assembled semi-automatically by merging airborne laser scans, building and road boundaries (including bridges) from national mapping data, and roof models derived by interactive stereo digitisation. The height accuracy is specified as $\pm$0.2$\,$m on buildings and $\pm$0.4$\,$m on general terrain. The model is based on data collected before 2015, and hence, differs from the state visible in the images in a handful of places. For evaluation, we thus ignore areas with obvious temporal differences due to construction activities. Note that the city model does not include vegetation. Consequently, ResDepth implicitly learns to filter out trees present in the initial model.

\newcommand{\ra}[1]{\renewcommand{\arraystretch}{#1}}
\begin{table*}[th]
	\centering
	\footnotesize
	\begin{adjustbox}{max width=\textwidth}
    		\begin{tabular}{@{}lccccccccc@{}}
    			\toprule
    			\textbf{} & \multicolumn{3}{c}{Overall} & \multicolumn{3}{c}{Building pixels} & \multicolumn{3}{c}{Terrain pixels w/o water}\\
    		\cmidrule(lr){2-4}\cmidrule(lr){5-7}\cmidrule(l{\tabcolsep}){8-10}
    			& MAE & RMSE & MedAE &  MAE & RMSE & MedAE & MAE & RMSE & MedAE \\
    			\midrule
    			Initial DEM                       & 2.81 & 4.49 & 1.43 & 2.44 & 3.93 & 1.33 & 3.08 & 4.81 & 1.58 \\
    			Median filtered                   & 2.75 & 4.41 & 1.39 & 2.39 & 3.87 & 1.31 & 3.00 & 4.72 & 1.53 \\
    			Unet-\textit{stereo}              & 7.79 & 9.31 & 6.77 & 8.69 & 10.14 & 8.20 & 7.43 & 9.04 & 6.34 \\
    			ResDepth-\textit{0}               & 1.43 & 2.85 & 0.61 & 2.14 & 3.95 & 0.85 & 1.01 & 1.95 & 0.53 \\
    			ResDepth-\textit{mono}            & 1.20 & 2.37 & 0.57 & 1.67 & 3.18 & 0.72 & 0.89 & 1.62 & 0.51 \\
    			ResDepth-\textit{stereo}          & 1.11 & 2.22 & 0.54 & 1.58 & 3.06 & 0.69 & 0.81 & 1.50 & 0.48 \\
    			ResDepth-\textit{stereo\textsubscript{iter}}     & 1.05 & 2.20 & 0.50 & 1.60 & 3.17 & 0.66 & 0.75 & 1.39 & 0.45 \\
    			\midrule
    			ResDepth-\textit{stereo} ($A$$\;\rightarrow\;$train, $B$$\;\rightarrow\;$test)          & 1.38 & 2.86 & 0.57 & 1.90 & 3.65 & 0.75 & 1.07 & 2.27 & 0.50 \\
    			ResDepth-\textit{stereo} ($B$$\;\rightarrow\;$train, $A$$\;\rightarrow\;$test)          & 1.30 & 2.68 & 0.56 & 1.96 & 3.71 & 0.77 & 0.91 & 1.78 & 0.49 \\
    			\bottomrule
    		\end{tabular}
	\end{adjustbox}
	\vspace*{\floatsep}
	\caption{Quantitative results on Zurich satellite data (in [m]). We evaluate the mean absolute error (MAE), the root mean square error (RMSE), and the median absolute error (MedAE). Residuals beyond $\pm$20$\,$m are discarded before computing statistics to account for temporal changes between ground truth and images. Building masks for object-specific metrics are dilated by 2 pixels (0.5$\,$m) to avoid aliasing at vertical walls. Results are averaged for ResDepth variants that are evaluated on multiple stereo pairs (last two rows).}
	\label{tab:ablation}
\end{table*}

\mypara{Network Training}
We vertically split the geographic area into five mutually exclusive stripes and use three stripes for training, one for validation, and one for testing. We use training patches with 128$\times$128 pixel dimensions (32$\times$32$\,$m in world coordinates). Terrain heights are globally normalised by centering to mean height~0 and scaled by the global standard deviation of the heights. To increase the amount of training data and to avoid biases due to the topography (sloped, south-facing terrain), we perform data augmentation by randomly rotating training patches with $\alpha\in\{90^\circ,180^\circ,270^\circ\}$ as well as horizontal and vertical flipping (see the extended report~\cite{stucker2020resdepth} for details).

We train the network in a fully supervised manner by minimising the pixel-wise absolute distance to ground truth depth maps (\ie, the $\ell_1$-loss). We use the adam optimizer~\cite{adam} with base learning rate $10^{-5}$, batch size 20, and weight decay of $10^{-5}$.

\subsection{ETH3D}
We also test ResDepth on close-range stereo pairs from the high-resolution multi-view ETH3D dataset~\cite{schoeps2017}. For the experiment, we downsample the images to a size of 886$\times$590 pixels, remove radial distortion, and convert them to grayscale. We limit ourselves to indoor scenes, and manually pick a set of binocular stereo pairs with high overlap and reasonable baselines. We use the PatchMatch stereo~\cite{bleyer2011patchmatch} implementation of COLMAP~\cite{schoenberger2016sfm,schoenberger2016mvs} to compute initial depths. Since there is no natural 2D coordinate system for ortho-rectification, we refine the depth map of the first image and warp the second image accordingly. 

\mypara{Network Training}
The data is split into training, validation, and test portions such that their fields of view are mutually exclusive, \ie, scene parts visible in the test set are never seen in the training or validation part. Because of the much larger depth range (relative to the baseline), we operate in inverse depth, \ie, pixel-wise depth values $d$ are converted to $1/d$ and scaled by the baseline for training and prediction. For the evaluation, we convert back to metric depth, as this is our target quantity. %
We train the ResDepth-\textit{stereo} network on patches of size 512$\times$512 cropped randomly from all training images. Gray-values are normalised to $[0\hdots 1]$ and scaled inverse depth values are centred to the mean of the patch. Patches are horizontally flipped at random for data augmentation. We again use adam, with base learning rate $10^{-5}$, batch size 4, and weight decay of $10^{-5}$.

\subsection{Results}
As error metrics, we use mean absolute error (MAE), root mean square error (RMSE), and median absolute error (MedAE). For the satellite data, we compute the error metrics also separately for buildings and terrain. Water bodies are excluded from the evaluation.

\def\rot#1{\rotatebox{90}{#1}}

\begin{figure*}[p]
  \centering
  \vspace{-0.5cm}
  \def\mywidth{0.24}
  \begin{tabular}{@{}c@{\hspace{1mm}}c@{\hspace{1mm}}c@{\hspace{1mm}}c@{\hspace{1mm}}c@{}}
    \small  & \small  & \small   & \small    \\
    \raisebox{.15\height}{\rot{\small  Reference view}}
    \hspace{0.05mm}
    \adjincludegraphics[width=\mywidth\linewidth,trim={0 0 0 0},clip]{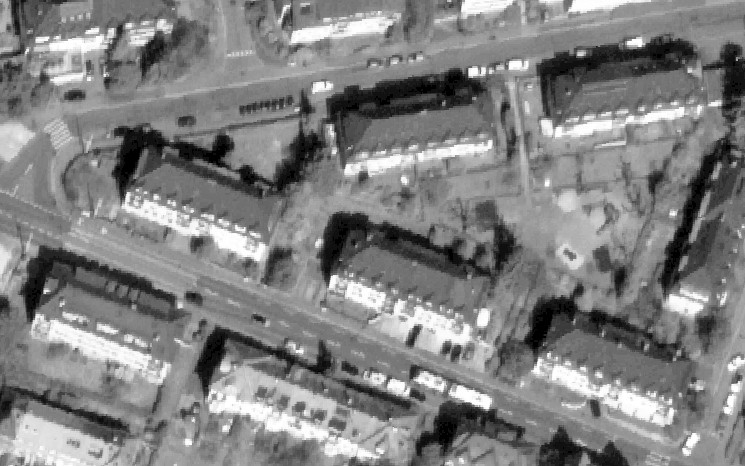} &
    \adjincludegraphics[width=\mywidth\linewidth,trim={0 0 0 0},clip]{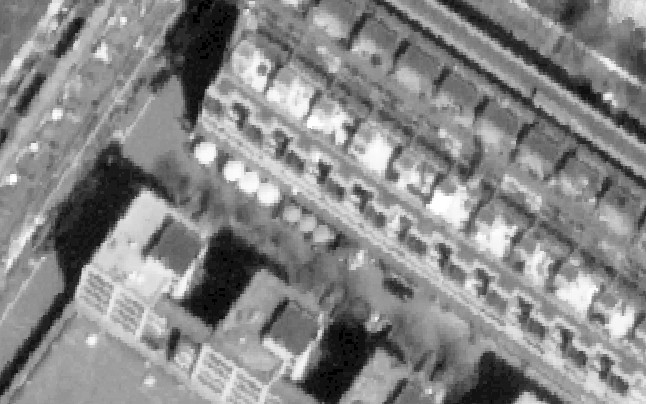} &
    \adjincludegraphics[width=\mywidth\linewidth,trim={0 0 0 0},clip]{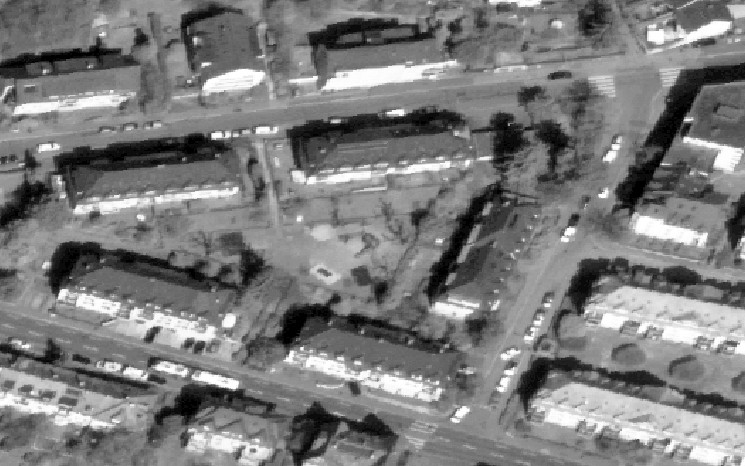} &
    \adjincludegraphics[width=\mywidth\linewidth,trim={0 0 0 0},clip]{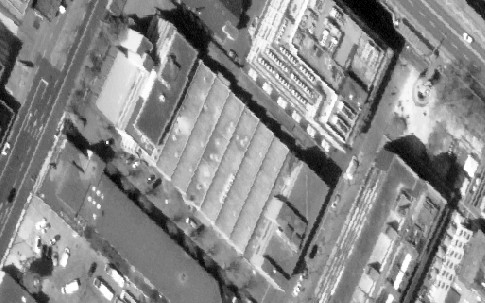}
    \\
    \raisebox{0.4\height}{\rot{\small Input DEM}}
    \adjincludegraphics[width=\mywidth\linewidth,trim={0 0 0 0},clip]{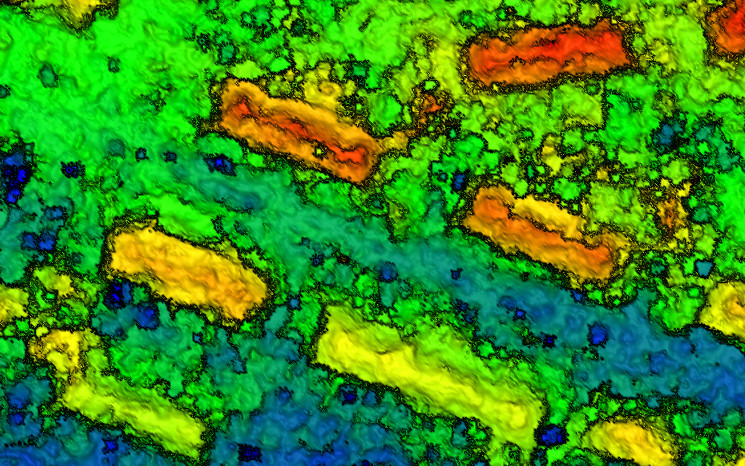} &
    \adjincludegraphics[width=\mywidth\linewidth,trim={0 0 0 0},clip]{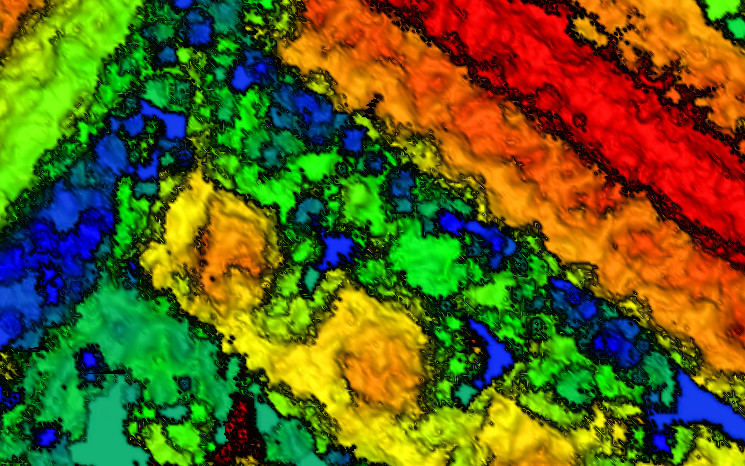} &
    \adjincludegraphics[width=\mywidth\linewidth,trim={0 0 0 0},clip]{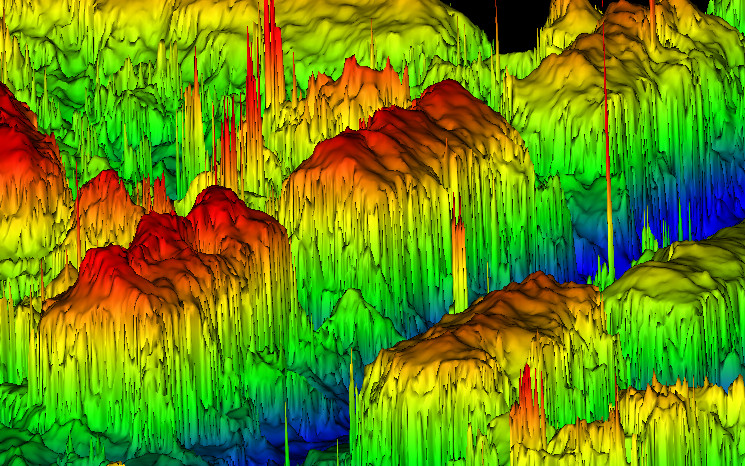} &
    \adjincludegraphics[width=\mywidth\linewidth,trim={0 0 0 0},clip]{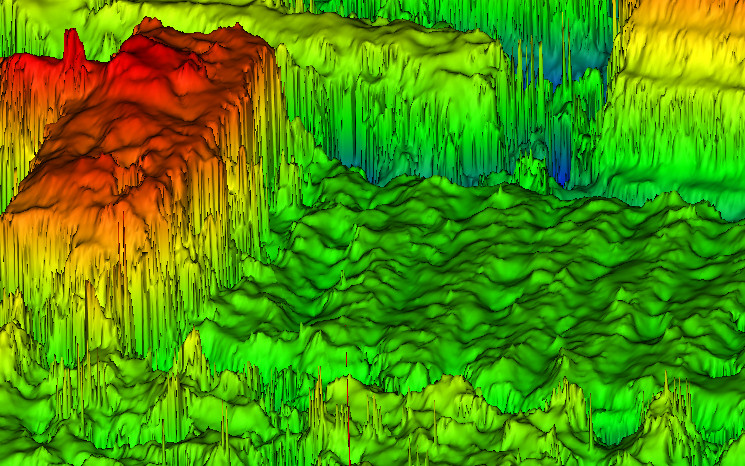} &
    \\
    \raisebox{.15\height}{\rot{\small Median filtered}}
    \hspace{0.05mm}
    \adjincludegraphics[width=\mywidth\linewidth,trim={0 0 0 0},clip]{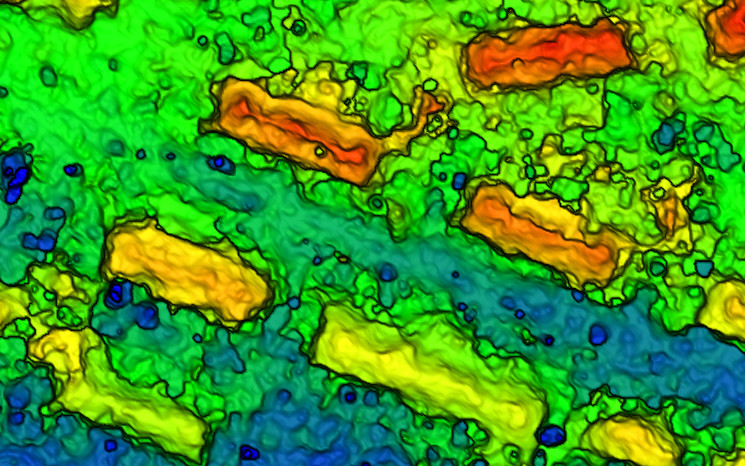} &
    \adjincludegraphics[width=\mywidth\linewidth,trim={0 0 0 0},clip]{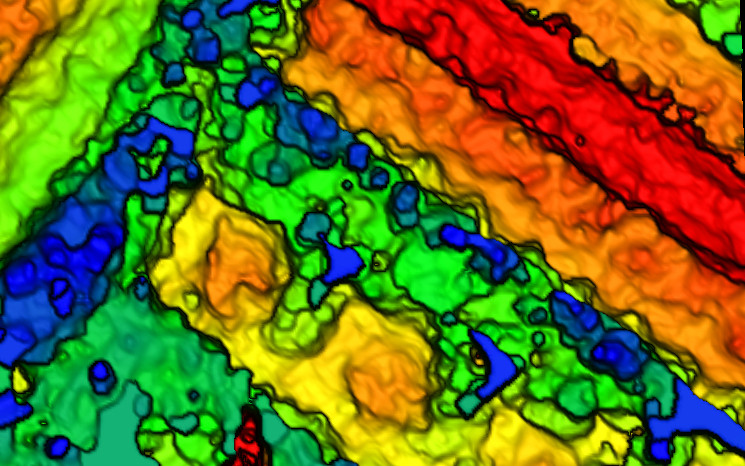} &
    \adjincludegraphics[width=\mywidth\linewidth,trim={0 0 0 0},clip]{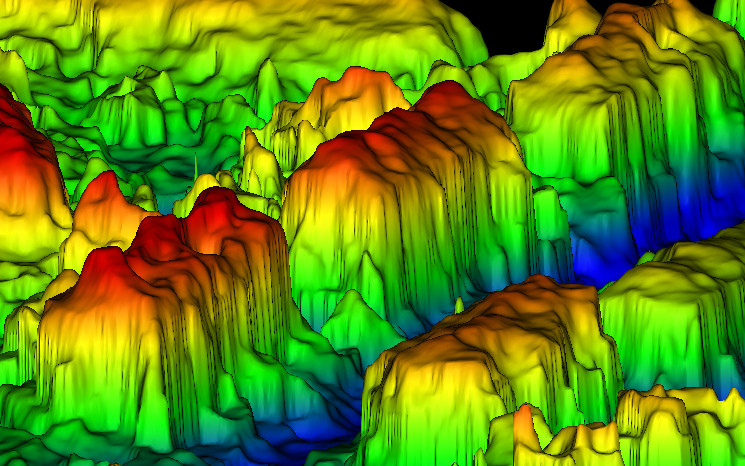} &
    \adjincludegraphics[width=\mywidth\linewidth,trim={0 0 0 0},clip]{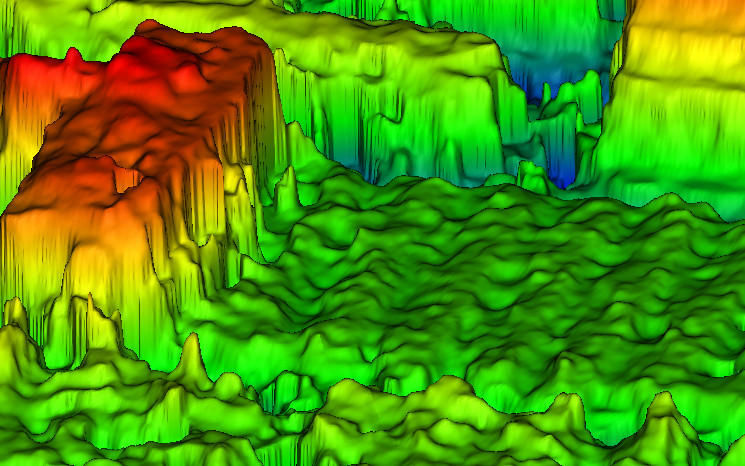} &
    \\
    \raisebox{.4\height}{\rot{\small ResDepth-\textit{0}}}
    \adjincludegraphics[width=\mywidth\linewidth,trim={0 0 0 0},clip]{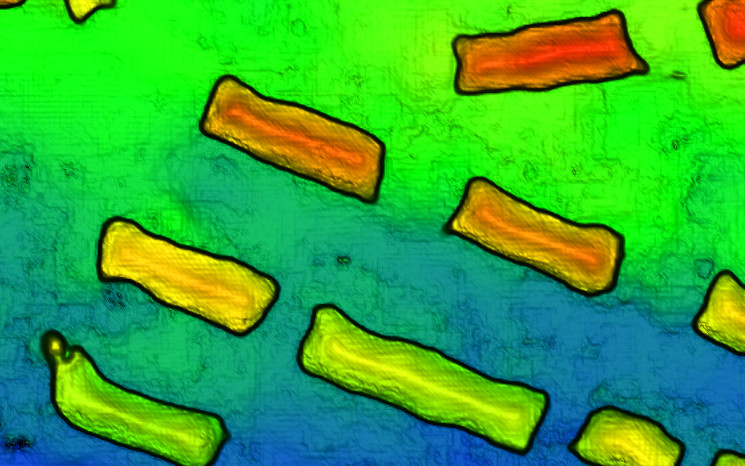} &
    \adjincludegraphics[width=\mywidth\linewidth,trim={0 0 0 0},clip]{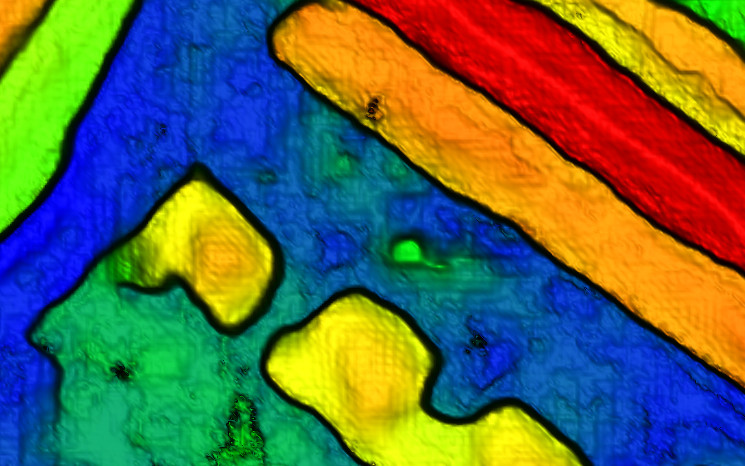} &
    \adjincludegraphics[width=\mywidth\linewidth,trim={0 0 0 0},clip]{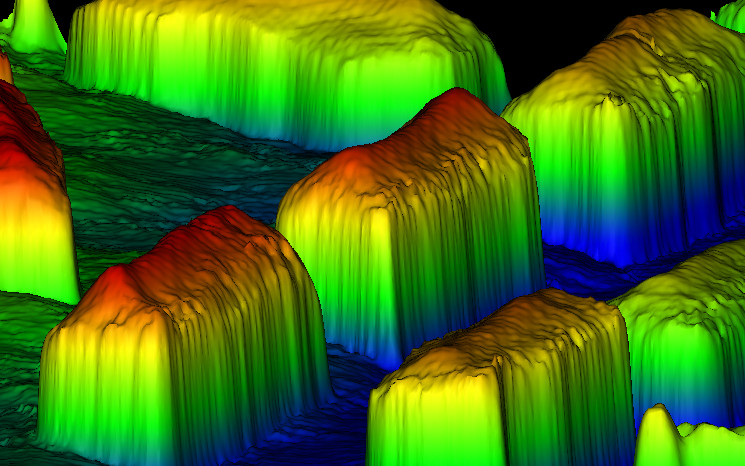} &
    \adjincludegraphics[width=\mywidth\linewidth,trim={0 0 0 0},clip]{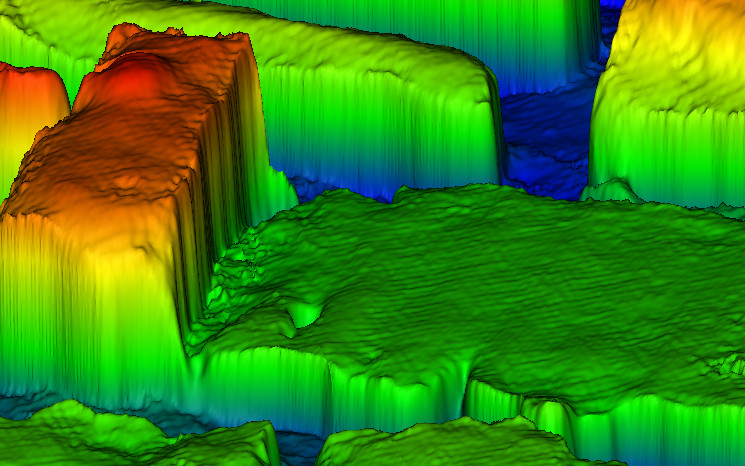} &
    \\
    \raisebox{0.15\height}{\rot{\small ResDepth-\textit{mono}}}
    \adjincludegraphics[width=\mywidth\linewidth,trim={0 0 0 0},clip]{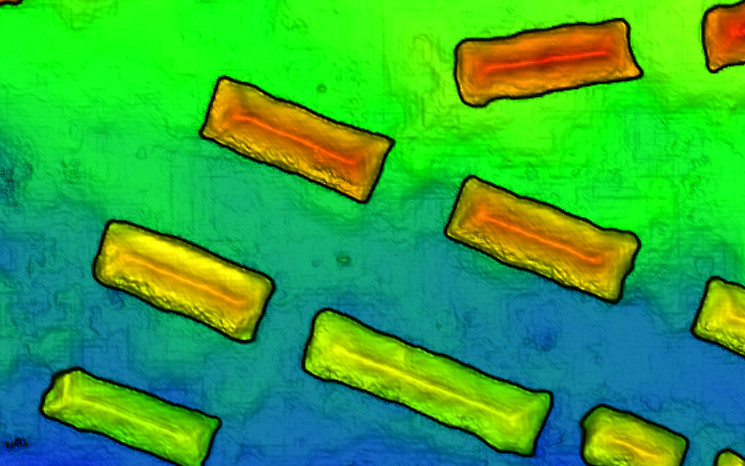} &
    \adjincludegraphics[width=\mywidth\linewidth,trim={0 0 0 0},clip]{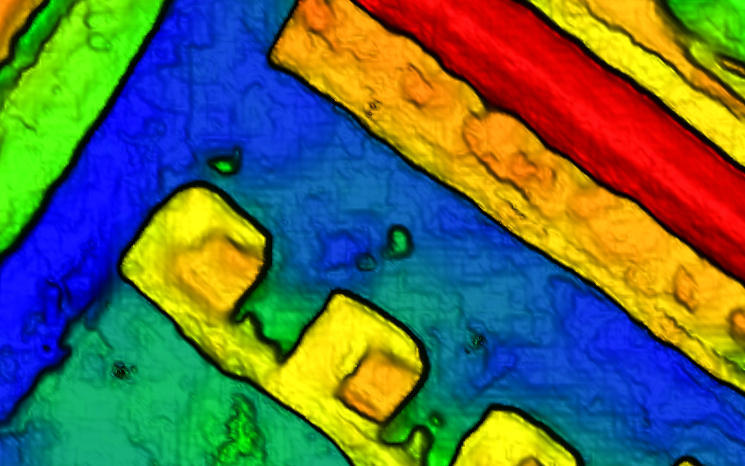} &
    \adjincludegraphics[width=\mywidth\linewidth,trim={0 0 0 0},clip]{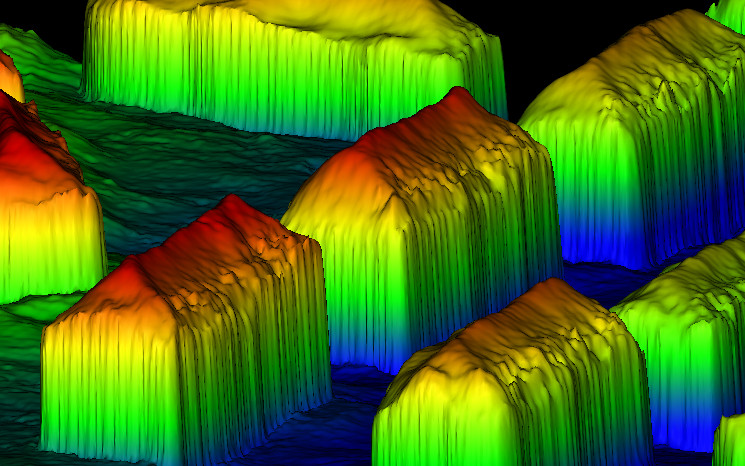} &
    \adjincludegraphics[width=\mywidth\linewidth,trim={0 0 0 0},clip]{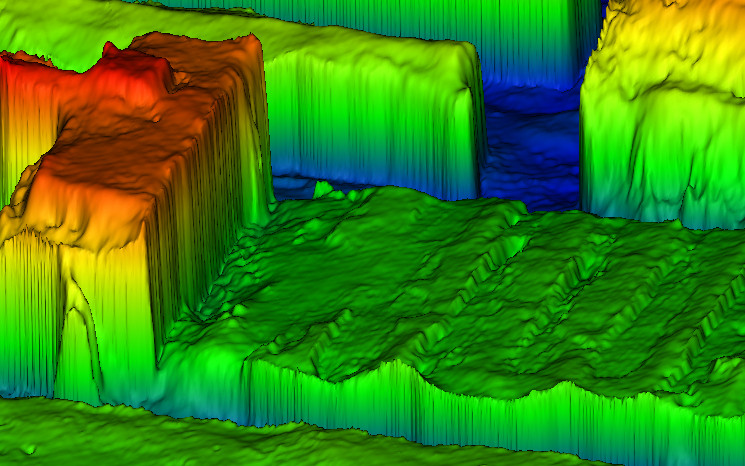} &
    \\
    \raisebox{0.15\height}{\rot{\small ResDepth-\textit{stereo}}}
    \adjincludegraphics[width=\mywidth\linewidth,trim={0 0 0 0},clip]{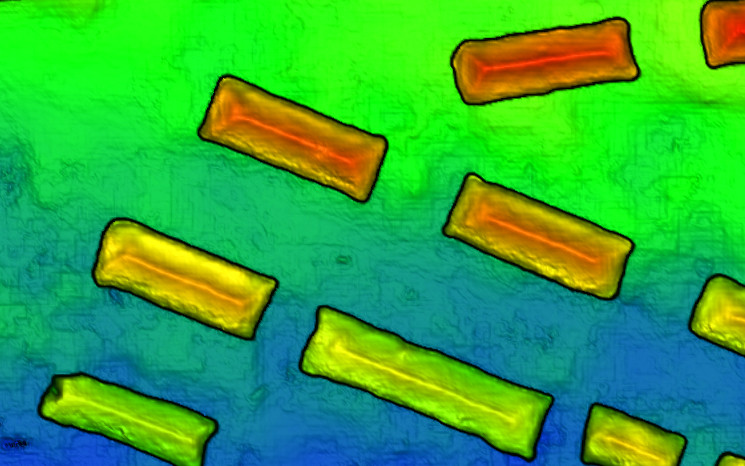} &
    \adjincludegraphics[width=\mywidth\linewidth,trim={0 0 0 0},clip]{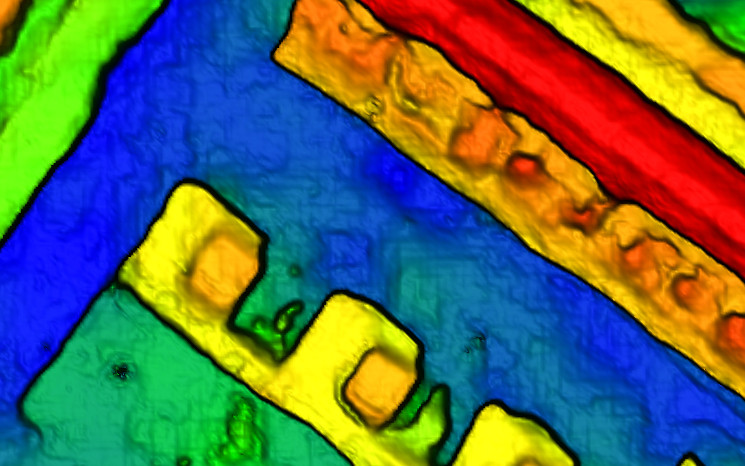} &
    \adjincludegraphics[width=\mywidth\linewidth,trim={0 0 0 0},clip]{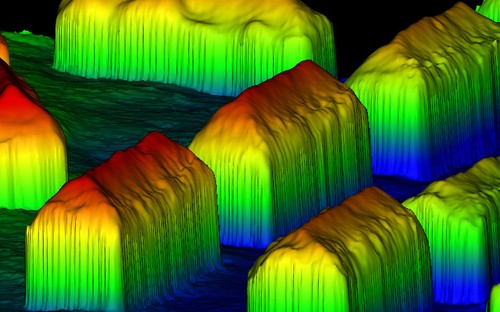} &
    \adjincludegraphics[width=\mywidth\linewidth,trim={0 0 0 0},clip]{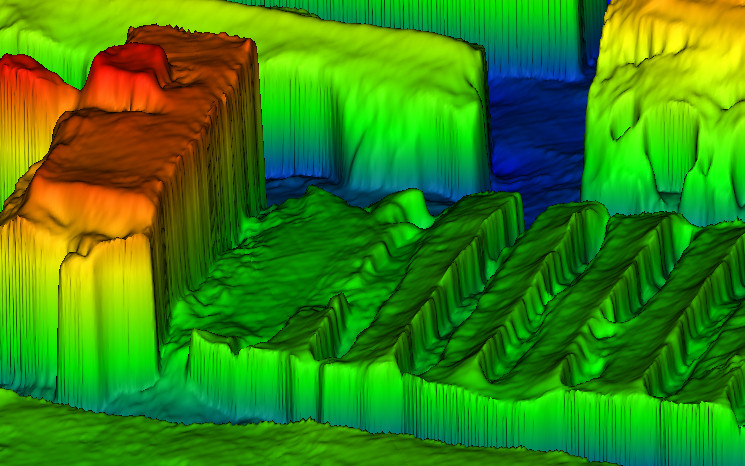} &
    \\
    \raisebox{.05\height}{\rot{\small ResDepth-\textit{stereo\textsubscript{iter}}}}
    \adjincludegraphics[width=\mywidth\linewidth,trim={0 0 0 0},clip]{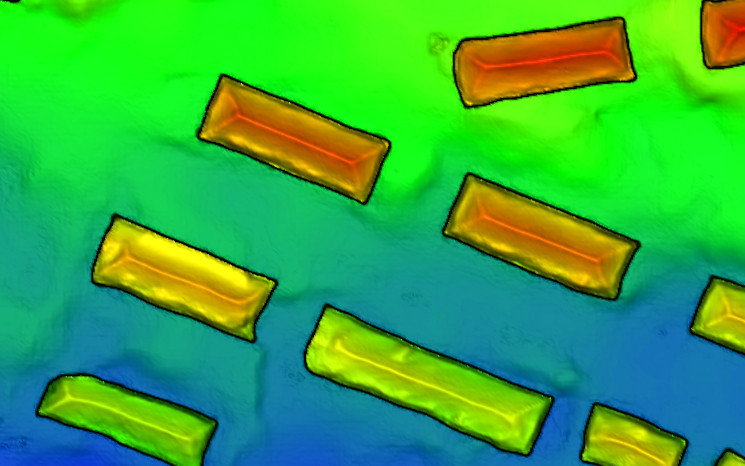} &
    \adjincludegraphics[width=\mywidth\linewidth,trim={0 0 0 0},clip]{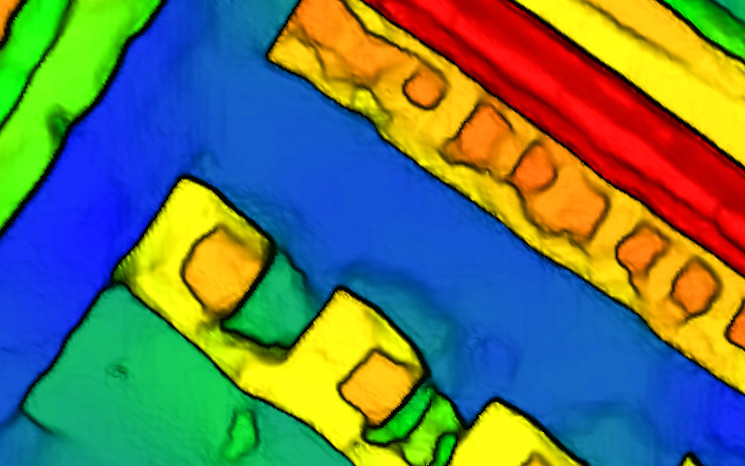} &
    \adjincludegraphics[width=\mywidth\linewidth,trim={0 0 0 0},clip]{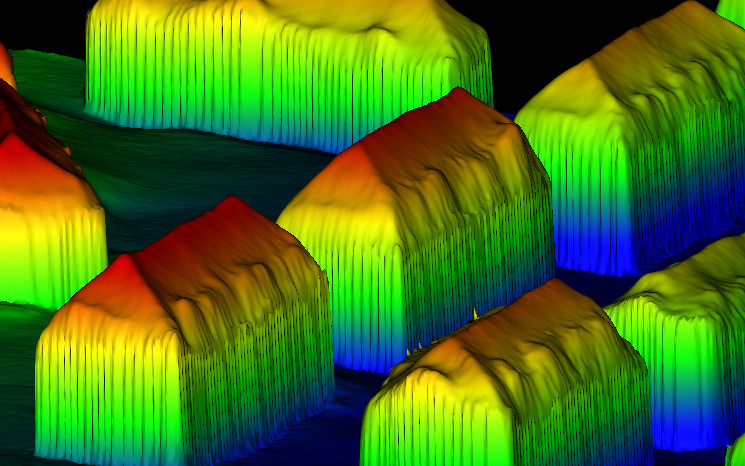} &
    \adjincludegraphics[width=\mywidth\linewidth,trim={0 0 0 0},clip]{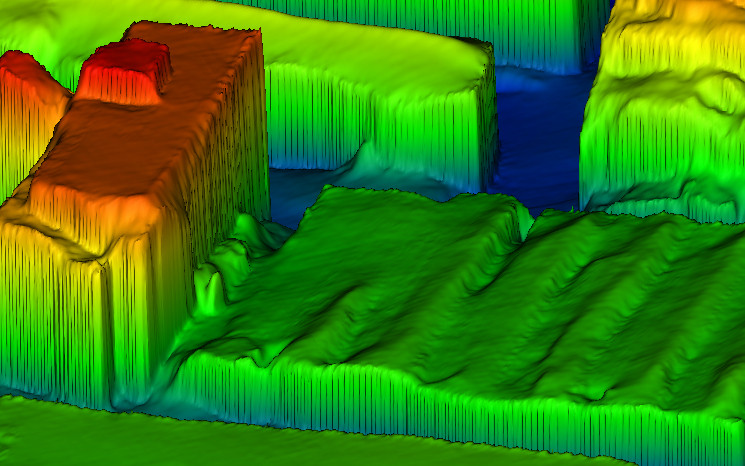} &
    \\
    \raisebox{.25\height}{\rot{\small Ground truth}}
    \hspace{0.05mm}
    \adjincludegraphics[width=\mywidth\linewidth,trim={0 0 0 0},clip]{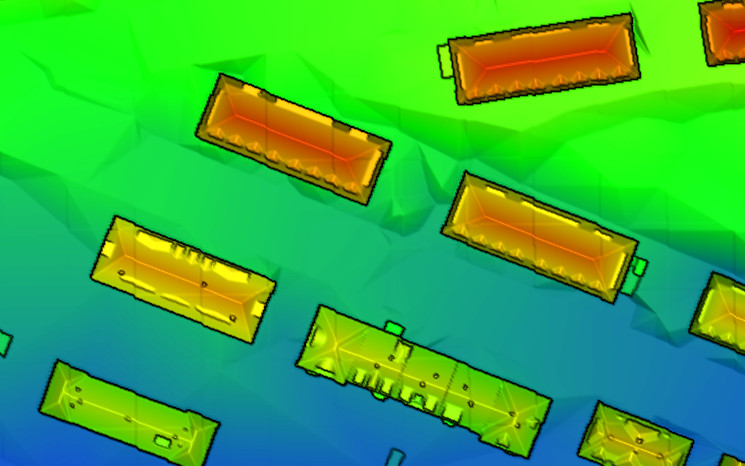} &
    \adjincludegraphics[width=\mywidth\linewidth,trim={0 0 0 0},clip]{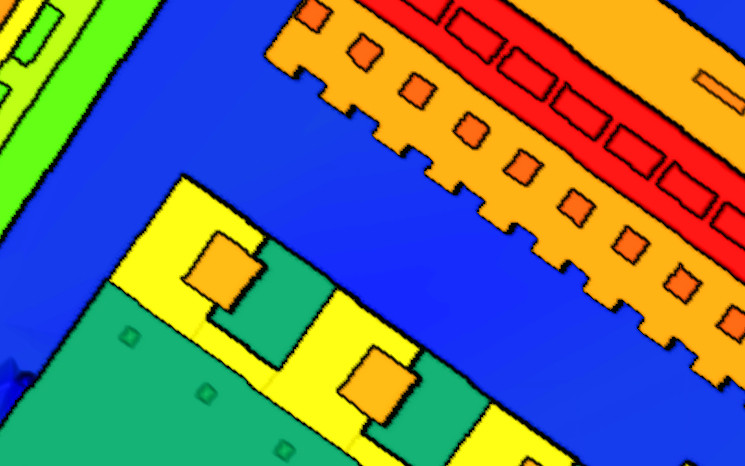} &
    \adjincludegraphics[width=\mywidth\linewidth,trim={0 0 0 0},clip]{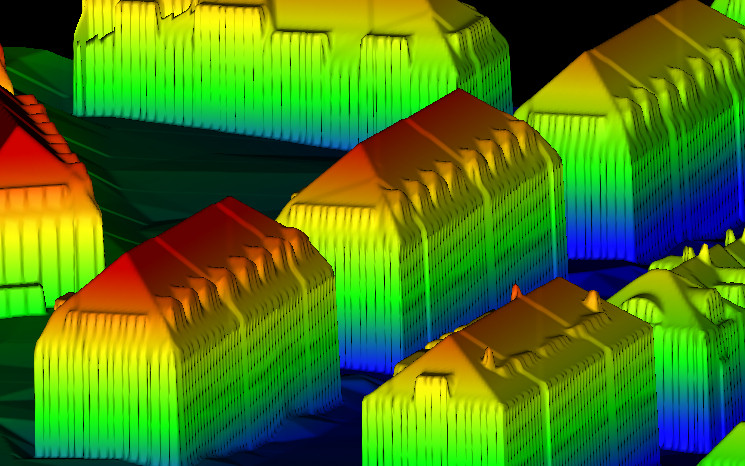} &
    \adjincludegraphics[width=\mywidth\linewidth,trim={0 0 0 0},clip]{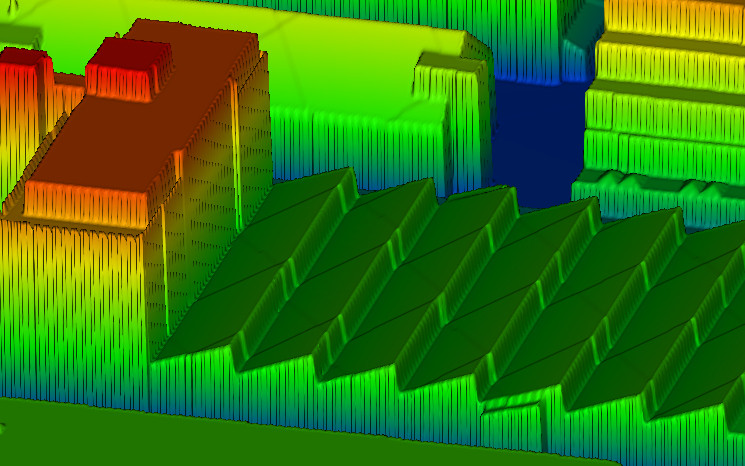} &
\end{tabular}
\vspace{1mm}
\caption{Visual comparison of different ResDepth variants over selected areas of Zurich. Heights are color-coded from blue to red.}
\label{fig:results_zurich}
\end{figure*}

\mypara{Ablation Study}
We first compare different DEM refinement strategies using a single satellite stereo pair. The initial DEM has a MAE of 2.81$\,$m and a MedAE of 1.43$\,$m, see Tab.~\ref{tab:ablation}. For completeness, we also show the RMSEs, which are between 1.5$\times$ and 2$\times$ higher than MAEs because of outliers but follow the same trend. Example regions from the DEM are shown in Fig.~\ref{fig:results_zurich}. The DEM is rather noisy. The comparatively high noise level is typical for satellite-derived models due to the limited sensor resolution, the large sensor-to-object distance, and the image quality.

As a baseline for a "cleaned" DEM, we follow a popular strategy implemented in several photogrammetry packages and apply a median filter (kernel size 5$\times$5) to denoise the DEM. This decreases the MAE only marginally and has little visual effect. The learned stereo baseline Unet-\textit{stereo} fails completely with a MAE of 7.8$\,$m. \Ie, with the same encoder-decoder architecture and the same amount of training data as used for ResDepth, it is not possible to learn full stereo matching end-to-end.

As expected, ResDepth-\textit{0} without any image evidence mostly acts as a context-aware, intelligent smoothing filter. Notably, it is already much better than the median filter baseline, reducing the MAE to 1.4$\,$m and the MedAE to 0.6$\,$m. Among others, it has learned the preference for vertical walls. Naturally, it is not able to add details missed by the original stereo matcher. Building outlines remain wobbly, and roofs are blobby or sagging (Fig.~\ref{fig:results_zurich}, 4$^\text{th}$ row).

ResDepth-\textit{mono} combines a coarse initial DEM with a single image, which works surprisingly well. Even without stereo observations, the correlation between intensity patterns in the image and depth patterns in the reconstruction contains a lot of useful information. Substructures on roofs emerge that were not visible in the input DEM, and straight lines and rectangular footprints are more faithfully reproduced (Fig.~\ref{fig:results_zurich}, 5$^\text{th}$ row). The MAE drops by 0.2$\,$m compared to \mbox{ResDepth-\textit{0}}. We speculate that also in other (learned or hand-crafted) pipelines that iteratively refine the reconstruction, a large portion of the improvement in later processing stages might come from this monocular "transfer" of crisp image structures, rather than from actual stereo correspondence.

\def\rot#1{\rotatebox{90}{#1}}

\begin{figure*}[!t]
  \centering
  \def\mywidth{0.19}
  \begin{tabular}{@{}c@{\hspace{1mm}}c@{\hspace{1mm}}c@{\hspace{1mm}}c@{\hspace{1mm}}c@{\hspace{1mm}}c@{}}
  \small Input view & \small Ground truth & \small COLMAP (unfiltered) & \small  ResDepth-\textit{stereo}  & \small ResDepth-\textit{stereo} (raw) & \    \\
  
  \adjincludegraphics[width=\mywidth\linewidth,trim={0 0 0 0},clip]{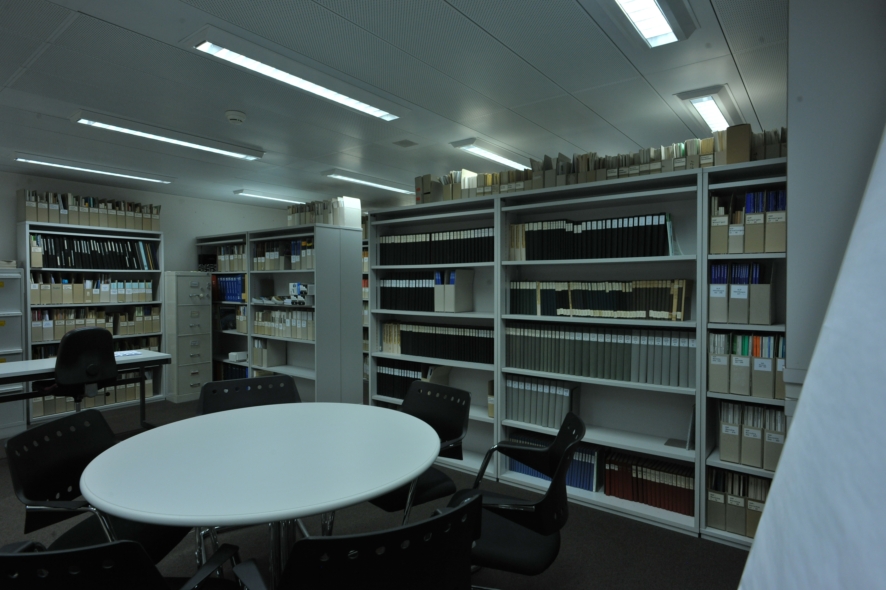} &
  \adjincludegraphics[width=\mywidth\linewidth,trim={0 0 0 0},clip]{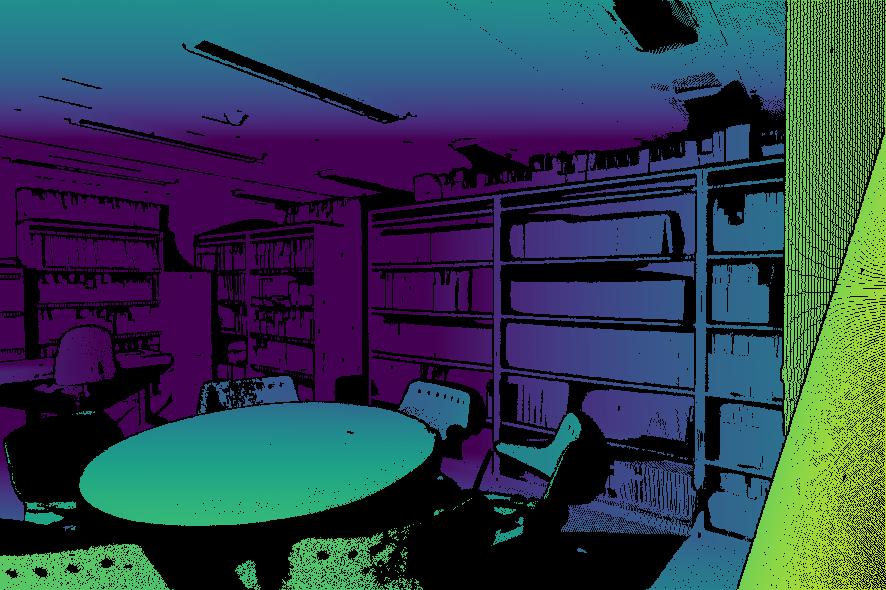} &
  \adjincludegraphics[width=\mywidth\linewidth,trim={0 0 0 0},clip]{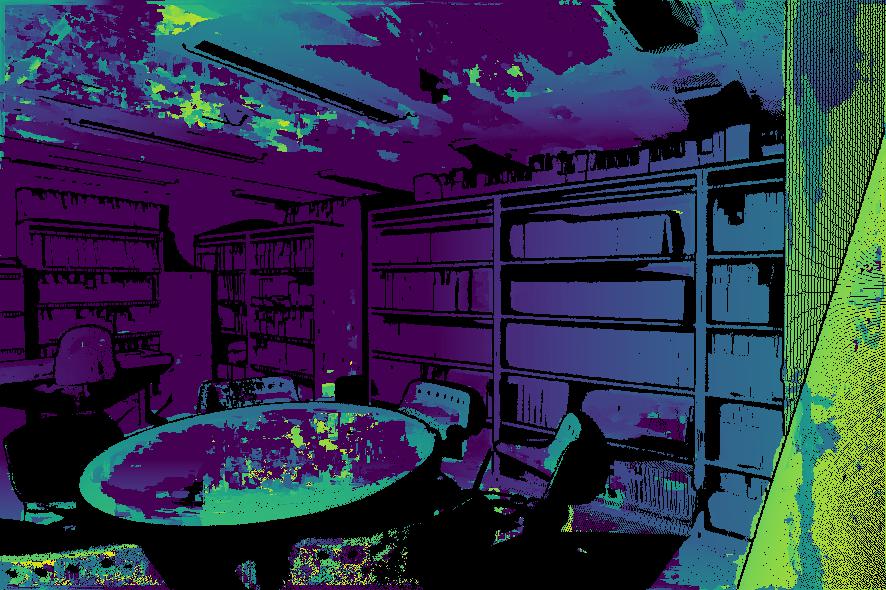} &
  \adjincludegraphics[width=\mywidth\linewidth,trim={0 0 0 0},clip]{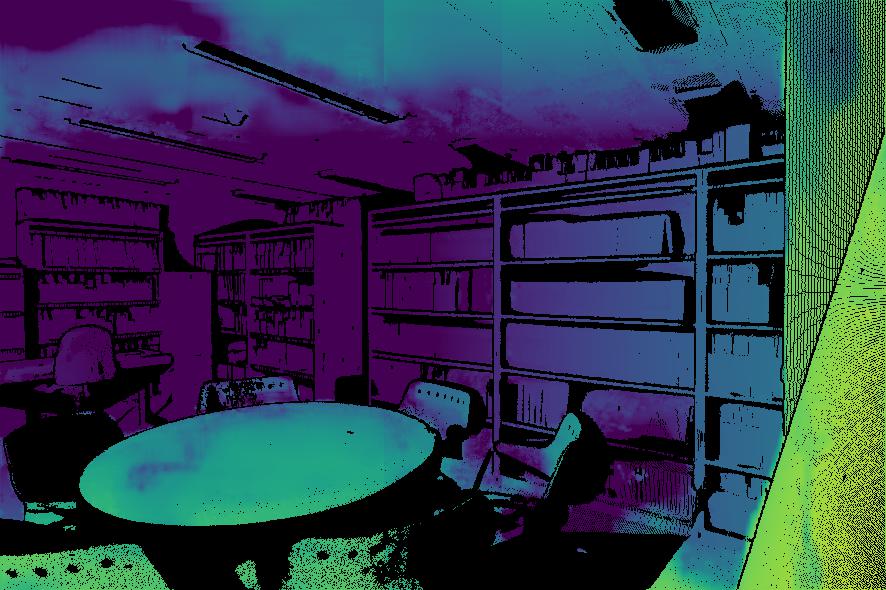} &
  \adjincludegraphics[width=\mywidth\linewidth,trim={0 0 0 0},clip]{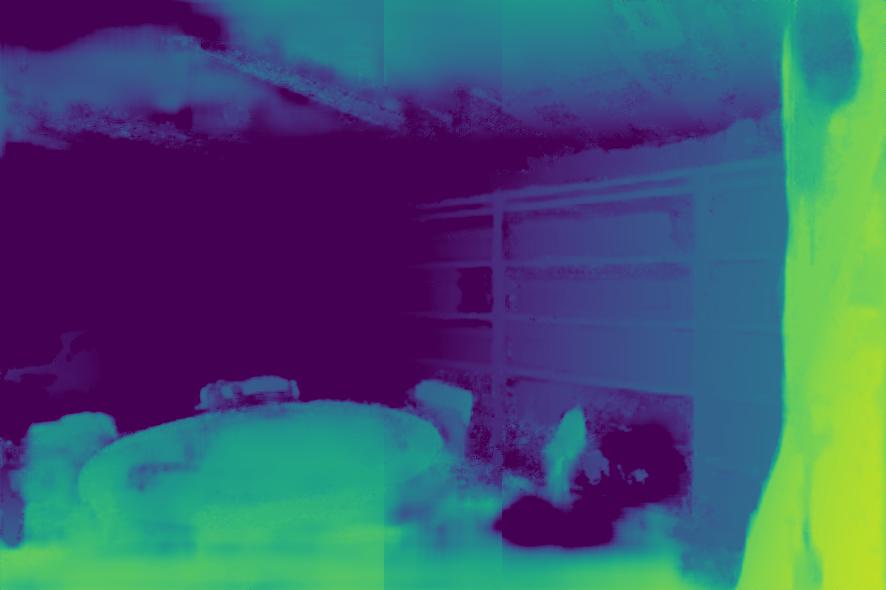}
  \\
  \adjincludegraphics[width=\mywidth\linewidth,trim={0 0 0 0},clip]{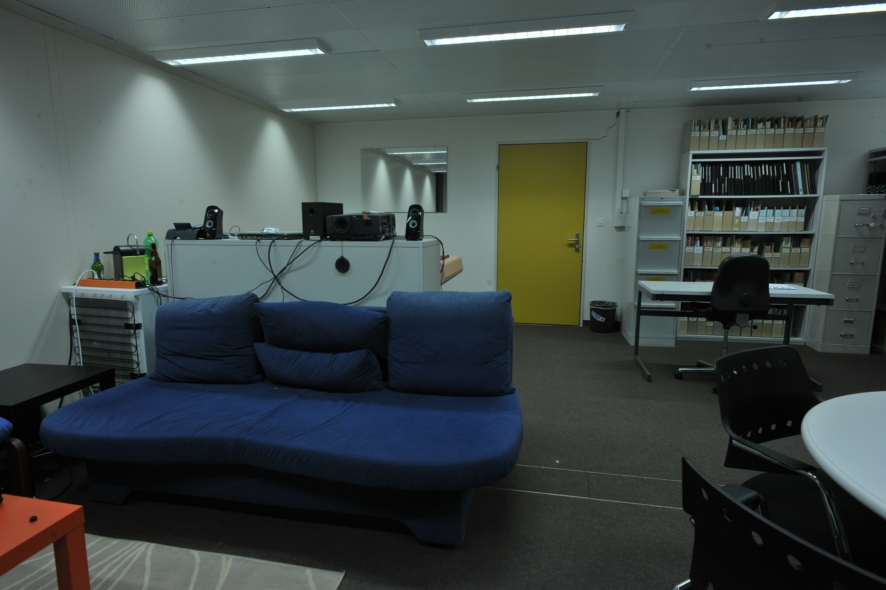} &
  \adjincludegraphics[width=\mywidth\linewidth,trim={0 0 0 0},clip]{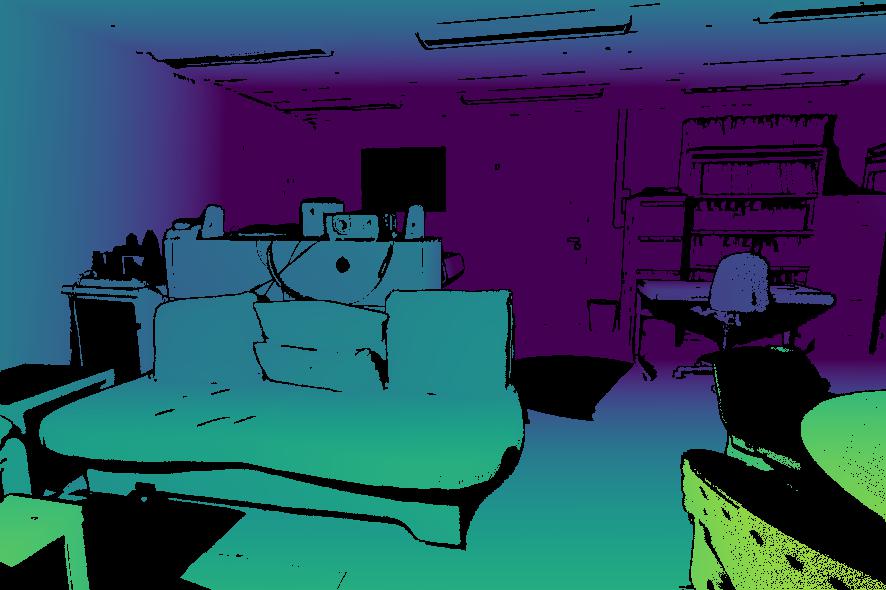} & \adjincludegraphics[width=\mywidth\linewidth,trim={0 0 0 0},clip]{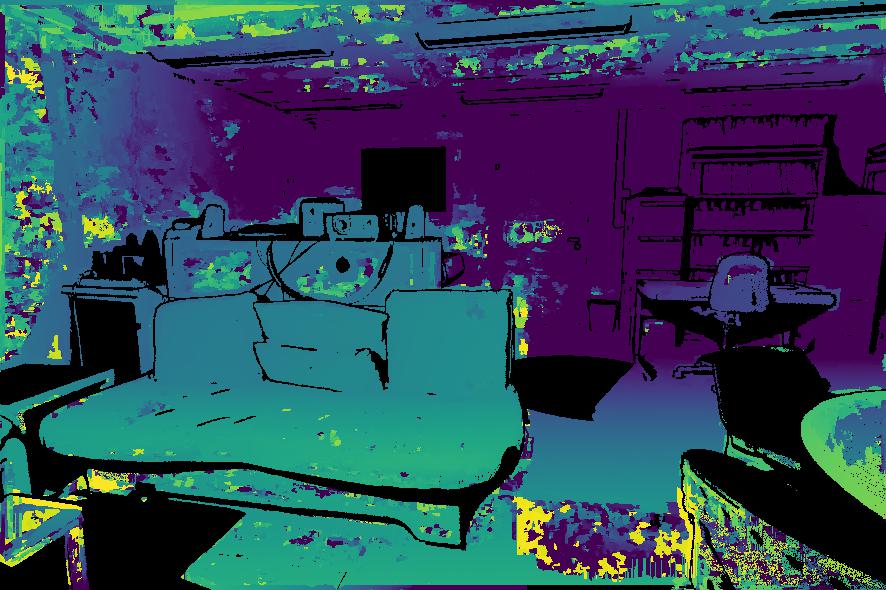} &
  \adjincludegraphics[width=\mywidth\linewidth,trim={0 0 0 0},clip]{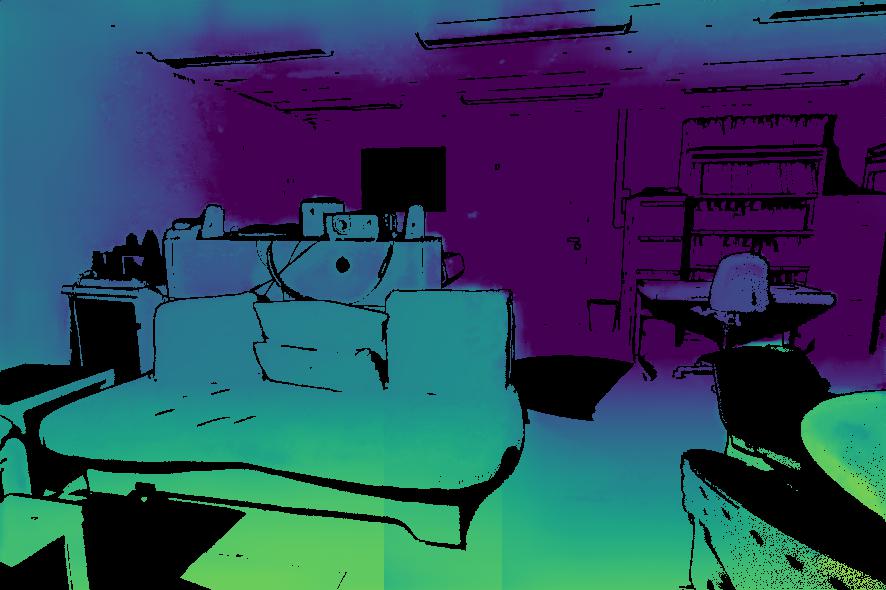} &
  \adjincludegraphics[width=\mywidth\linewidth,trim={0 0 0 0},clip]{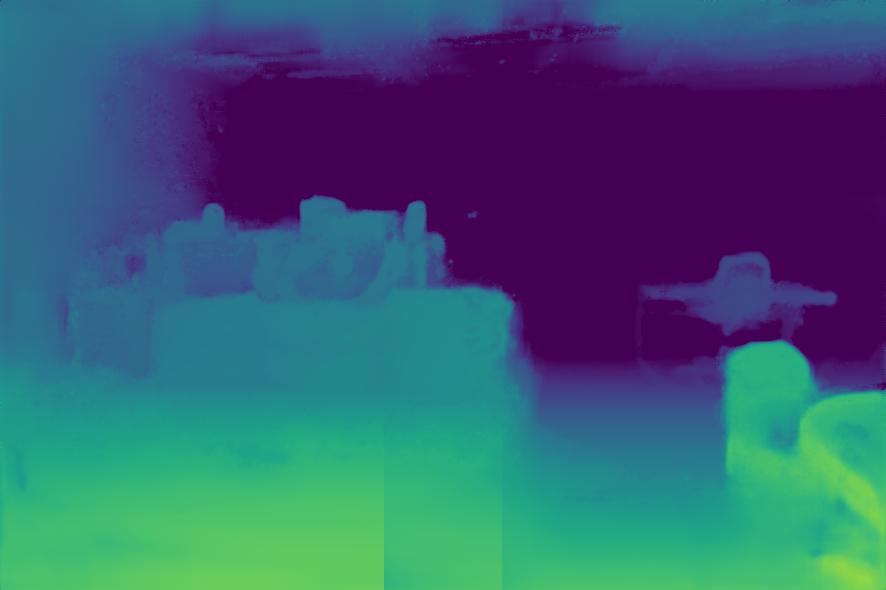}
\end{tabular}
\vspace{0.5em}
\caption{Example results for ETH3D indoor scenes. Pixels without valid ground truth depth are displayed in black (except in the rightmost column, which shows the unmasked ResDepth prediction). For reference, we compare the unmasked maps produced by ResDepth (rightmost column) to the dense matcher of COLMAP with heuristic filtering turned off to obtain dense depth maps.}
\label{fig:results_eth3d}
\end{figure*}

ResDepth-\textit{stereo} further improves the reconstruction. Visually, the difference is, in fact, larger than suggested by the quantitative improvement of 0.1$\,$m in MAE, as building shapes become crisper and additional roof details emerge (Fig.~\ref{fig:results_zurich}, 6$^\text{th}$ row). The iterative optimisation of \mbox{ResDepth-\textit{stereo\textsubscript{iter}}} yields only small quantitative gains but visibly sharper and more detailed 3D geometry (Fig.~\ref{fig:results_zurich}, 7$^\text{th}$ row). Even small dormers are discernible and spurious bumps on roads disappear. Uncommon building structures are sometimes over-smoothed (\eg, the saw-tooth roof in Fig.~\ref{fig:results_zurich}, 4$^\text{th}$ column), probably because they are not represented in the training data. The final MedAE of our best result after two rounds of ResDepth is 0.5$\,$m, the MAE is roughly 1.0$\,$m. These values are quite remarkable given the image resolution of $\approx$ 0.5$\,$m and the uncertainty of the satellite poses on the order of 0.5$\,$m on the ground. 

\mypara{Generalisation Across Images}
In practice, the train and test regions will be seen in different images, so the model must also generalise across changes in lighting and viewing directions. When applied in that scenario, ResDepth-\textit{stereo} pays only a small performance penalty and still improves significantly over the input DEM (Tab.~\ref{tab:ablation}, last two rows). The overall MAE increases by 0.2 to 1.3$\,$m. We note that these results, while using different, separate satellite images, are still for the city of Zurich and are expected to deteriorate as one moves to other geographic locations with different urban planning and architectural style. Yet, given enough training data, it seems realistic to learn a model that generalises across locations and variations of urban style.

\mypara{Generalisation to Other City Regions} 
We apply ResDepth-\textit{stereo} without further fine-tuning to geographically more distant parts of Zurich to quantify how well ResDepth generalises across a city. Due to space restrictions, we provide these results in the extended report~\cite{stucker2020resdepth}.

\mypara{ETH3D}
We show quantitative results in Tab.~\ref{tab:eth3d} and visual examples in Fig.~\ref{fig:results_eth3d}. We follow the same evaluation protocol as~\cite{huang2018deepmvs} and compare our dense, refined depth maps with unfiltered dense depth maps of COLMAP. This comparison should be taken with a grain of salt: most of the error of COLMAP is due to areas that are masked out if its internal filter is switched on. \Ie, COLMAP is aware of the problems and would discard them if permitted, at the cost of lower completeness.
ResDepth consistently improves over the COLMAP baseline, in some scenes considerably. On average, the MAE decreases from 0.35$\,$m to 0.15$\,$m. While those numbers are very encouraging, the visual impact of ResDepth is smaller in the close-range scenario. In the urban modelling case, the images and the ground truth DEM contain a lot of structures that it can learn to exploit, like building shapes, rooflines, trees, \etc. Objects in the close-range images are much larger (relative to the pixel size), there are few discontinuities, and even over fairly large neighbourhoods, the prevalent structure is planarity. ResDepth does manage to correct quite a bit of detail w.r.t.\ the COLMAP input, \eg, the chairs in the 1$^\text{st}$ row and the area around the sofa in the 2$^\text{nd}$ row in Fig.~\ref{fig:results_eth3d}. But the main driver of its good performance appears to be that it has learned the sensible, yet unspectacular prior to associate featureless, homogeneous image regions with planar surfaces (\eg, the table in the 1$^\text{st}$ row and the floor in the 2$^\text{nd}$ row in Fig.~\ref{fig:results_eth3d}).

\begin{table}[t]
	\centering
	\footnotesize
	\begin{adjustbox}{max width=\columnwidth}
    	\ra{1.1}  %
    		\begin{tabular}{@{}lccccc@{}}
    			\toprule
    			 & \multirow{2}{*}{\parbox{1cm}{Depth-stereo\\ pairs}} & \multicolumn{2}{c}{ResDepth-\textit{stereo}} & \multicolumn{2}{c}{COLMAP (unfilt.)}\\
    			\cmidrule(lr){3-4}\cmidrule(lr){5-6}
    			& & MAE & RMSE  & MAE & RMSE\\
    			\midrule
    			Delivery area & 4 & 0.21 & 0.75 & 0.24 & 0.94 \\
    			Kicker        & 4 & 0.16 & 0.42 & 0.41 & 1.11 \\
    			Office        & 3 & 0.13 & 0.22 & 0.55 & 1.21 \\
    			Pipes         & 3 & 0.19 & 0.60 & 0.45 & 1.30 \\
    			Relief        & 4 & 0.18 & 0.85 & 0.28 & 1.11 \\
    			Relief\_2      & 4 & 0.12 & 0.57 & 0.17 & 0.82 \\
    			Terrains      & 4 & 0.07 & 0.20 & 0.45 & 1.42 \\
    			\midrule
    			Overall       & 26 & 0.15 & 0.57 & 0.35 & 1.13  \\
    			\bottomrule
    		\end{tabular}
	\end{adjustbox}
	\vspace*{\floatsep}
	\caption{Quantitative results on ETH3D indoor scenes (in [m]). Residuals beyond $\pm$10$\,$m are discarded before computing statistics, since they usually occur only outside of the stereo overlap.}
	\label{tab:eth3d}
\end{table}
\section{Conclusion}
We have presented an astonishingly simple, yet highly effective way to use deep networks for dense 3D reconstruction: instead of replacing existing (hand-crafted or learned) stereo matchers with a deep network, we complement them. The network receives as input both an initial DEM and the stereo images and estimates a depth correction that is added to the DEM to improve it. We have shown that this strategy can reduce the errors of state-of-the-art stereo matchers more than 2$\times$, with a standard Unet architecture without any special modifications and a moderate amount of training data. The computational cost is one forward-pass per tile, with tile size depending on GPU memory. Inference for our test area takes less than 2~sec. The main burden is offline ortho-rectification.

In future work, we plan to test the ResDepth idea also for multi-view stereo, which is straightforward since all views are warped to the same image coordinate system. Furthermore, it may be possible to train ResDepth in such a way that it gradually refines the reconstruction over multiple iterations, with the same set of weights.

At a meta-level, we see ResDepth as a reminder to keep things simple and a strong baseline that should not be overlooked when designing more sophisticated stereo networks.

{\small
\bibliographystyle{ieee_fullname}
\bibliography{egbib}
}

\clearpage
\newpage
\appendix
\renewcommand{\thefigure}{A\arabic{figure}}
\renewcommand{\theHfigure}{A\arabic{figure}}
\setcounter{figure}{0}

\renewcommand{\thetable}{A\arabic{table}}
\renewcommand{\theHtable}{A\arabic{table}}
\setcounter{table}{0}

\captionsetup[subfloat]{captionskip=7pt, farskip=10pt}
\renewcommand{\thefigure}{\thesection.\arabic{figure}}
\renewcommand{\thetable}{\thesection.\arabic{table}}

\twocolumn[
    \begin{center}
        \vskip .375in
        {\Large \bf ResDepth: Learned Residual Stereo Reconstruction \par}
        \vskip .5em
        {\large \bf Supplementary Material \par}
        \vspace*{24pt}
        { 
          \large
          \lineskip .5em
          \begin{tabular}[t]{c}
             Corinne Stucker\qquad Konrad Schindler \vspace{1mm}\\
            Photogrammetry and Remote Sensing, ETH Zurich, Switzerland\\
            {\tt\small \{firstname.lastname@geod.baug.ethz.ch\}}
            \vspace*{1pt}
          \end{tabular}
          \par
          }
        \vskip .5em
        \vspace*{12pt}
    \end{center}
]

\def\thesection{\Alph{section}}
In this document, we provide additional details, complementary visualisations, and further experiments that we could not include in the main paper due to space restrictions. In particular,

\begin{itemize}
  \item we show the full extent of the Zurich dataset and mention further technical details about the model training.
  \item we illustrate the effect of rectifying satellite images with the initial surface model.
  \item we present additional results on large-scale urban DEM refinement. We apply ResDepth-\textit{stereo} without further fine-tuning to geographically more distant regions to assess if the model learns generic features that generalise to unseen locations, albeit still for the city of Zurich. Furthermore, we interchange the mutually exclusive groups of stereo pairs used during training and inference to validate if our separation into training and test stereo pairs affects the model performance.
  \item we demonstrate that ResDepth learns a generic prior for DEM refinement that can be used to refine any DEM, irrespective of which method was used to generate it. \Ie, during training, one only has to supply initial DEMs generated by one's favourite stereo matcher, and the network will implicitly learn to adapt to the specific biases and error patterns induced by the initial stereo method. We qualitatively show ResDepth's ability to seamlessly complement any stereo matching pipeline by training two network variants: the first network is trained to refine DEMs obtained with hierarchical semi-global matching~\cite{rothermel2012sure},\footnote{As in the main paper.} whereas the second network is tailored to DEMs derived from the publicly available \textit{s2p} satellite stereo pipeline \citeSupp{deFranchis2014automatic}.
\end{itemize}

\newpage

\section{Zurich DEM Dataset}
We evaluate our method on satellite images acquired over Zurich, Switzerland. The study area is located to the northwest of the city centre and covers $\approx$2.3$\,$km$^\text{2}$ (see Fig.~\ref{fig:areas}, yellow rectangle). It includes several urban construction styles, ranging from widely spaced, detached residential buildings to allotments and tall commercial buildings. Furthermore, it covers flat and rather steep terrain, a forested hill, and a stretch of the river Limmat.

\begin{figure*}[t!]
\centering
\includegraphics[width=\textwidth]{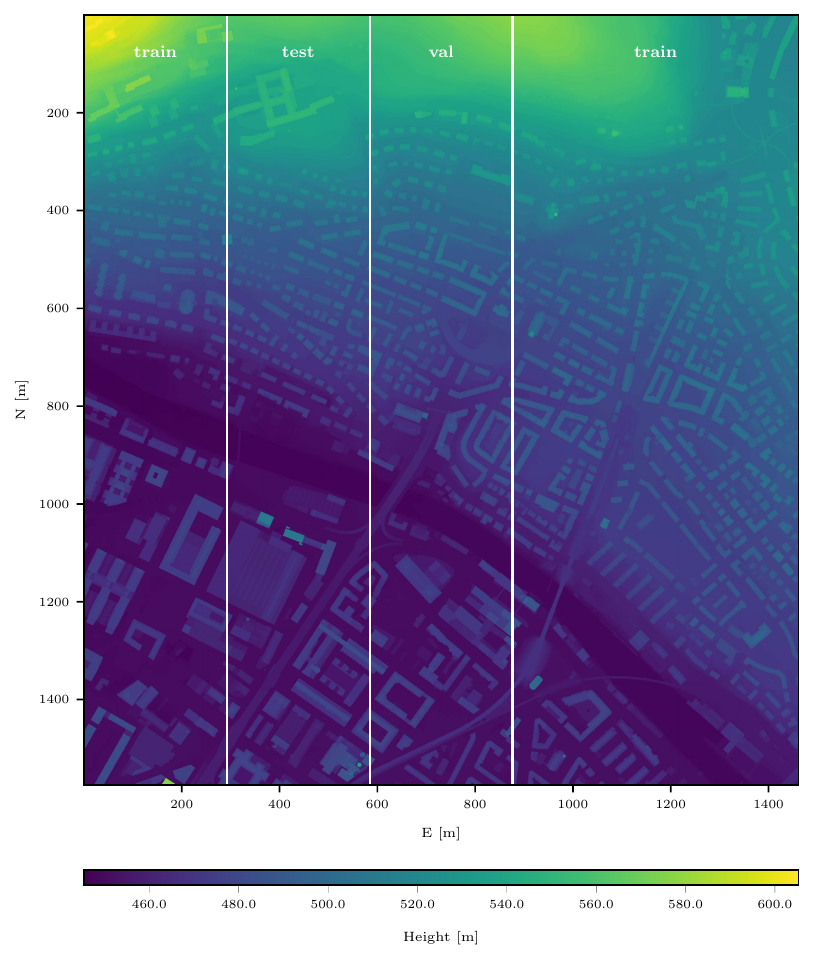}
\caption{Ground truth DEM of Zurich, separated into training, validation, and test region. The axes show relative coordinates w.r.t.\ to the upper left corner of the raster DEM. The coordinates of the upper left corner are (463395.0, 5249777.0), given in the local UTM 32N coordinate system. The colors indicate ellipsoidal heights w.r.t. the GRS80 ellipsoid.}
\label{fig:data_allocation}
\end{figure*}

We use state-of-the-art hierarchical semi-global matching~\cite{rothermel2012sure} tailored to satellite images to generate the initial DEM of that area. The grid spacing is 0.25$\,$m, corresponding to a total of $\approx 3.7\cdot 10^7$ pixels. Fig.~\ref{fig:data_allocation} depicts the corresponding ground truth DEM rendered from the official city model of Zurich.\footnote{\url{https://www.stadt-zuerich.ch/ted/de/index/geoz/geodaten_u_plaene/3d_stadtmodell.html}}

We vertically split the area into five mutually exclusive stripes and use three stripes for training, one for validation, and one for testing. The test stripe was chosen to include the full range of buildings (small and large, low and high, residential and industrial) and to avoid regions where the ground truth model is inaccurate, particularly around complex bridge structures.

\section{Image Rectification}
\begin{figure*}[!t]
  \centering
  \def\mywidth{0.22}
  \begin{tabular}{@{}c@{\hspace{2mm}}c@{\hspace{2mm}}c@{\hspace{2mm}}c@{\hspace{1mm}}c@{}}
  \small \nth{1} stereo image & \small \nth{2} stereo image & \small \nth{1} stereo image rectified & \small \nth{2} stereo image rectified \vspace{1mm}\\
  \adjincludegraphics[width=\mywidth\linewidth,trim={0 0 0 0},clip]{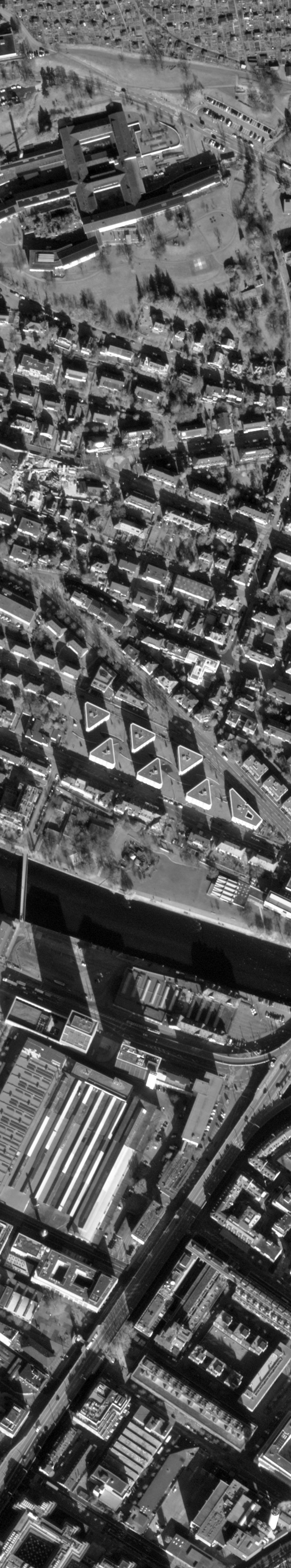} &
  \adjincludegraphics[width=\mywidth\linewidth,trim={0 0 0 0},clip]{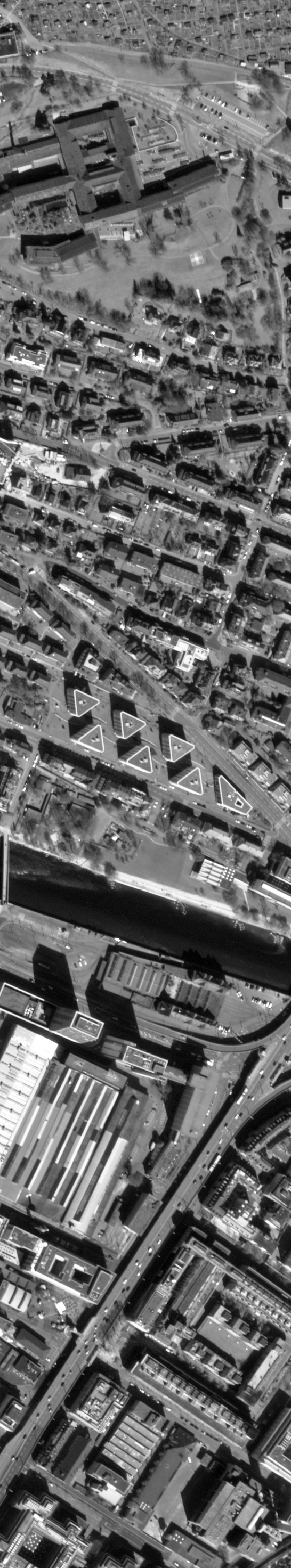} &
  \adjincludegraphics[width=\mywidth\linewidth,trim={0 0 0 0},clip]{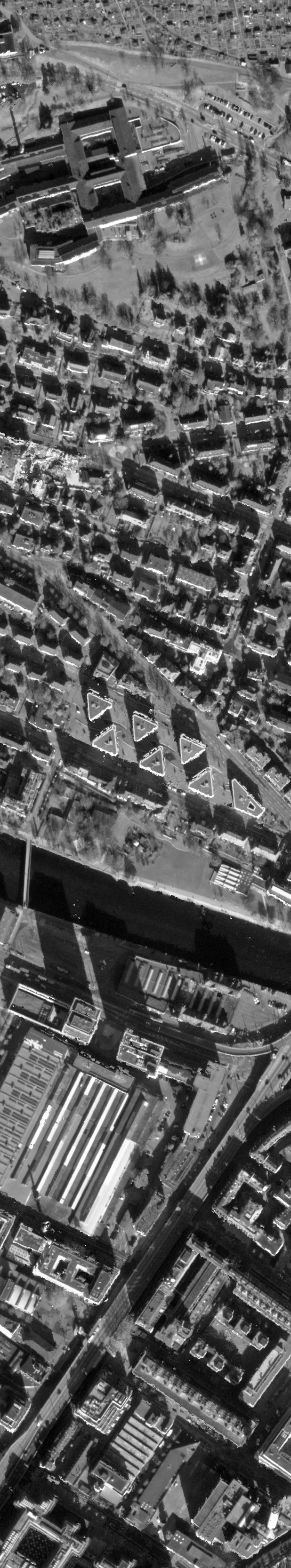} &
  \adjincludegraphics[width=\mywidth\linewidth,trim={0 0 0 0},clip]{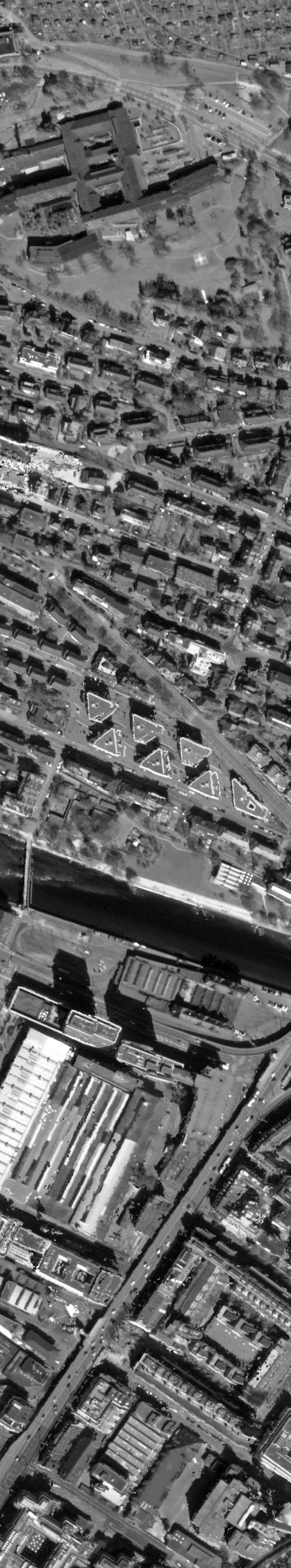} 
  
\end{tabular}
\vspace{0.5em}
\caption{Satellite stereo pair displaying the test region of Zurich. The \nth{1} and \nth{2} column show the raw satellite images, the \nth{3} and \nth{4} column are the ortho-rectified stereo pairs used by the ResDepth network.}
\label{fig:zurich_orthos}
\end{figure*}

We show the satellite images used for refining the test region of the Zurich DEM in Fig.~\ref{fig:zurich_orthos}, \nth{1} and \nth{2} column. The ortho-rectified counterparts are displayed in the \nth{3} and \nth{4} column. Note that we deliberately do not perform any ray-casting to account for occlusions during the ortho-rectification process. Instead, we prefer to render duplicate textures if the corresponding rays intersect the surface twice, leading to photometrically inconsistent, systematically displaced copies of nearby textures. This effect is particularly visible for tall buildings.

\clearpage

\def\rot#1{\rotatebox{90}{#1}}

\clearpage

\section{Additional Results}
\subsection{Geographical Generalisation within Zurich}

\begin{figure}[ht]
\centering
\includegraphics[width=\columnwidth]{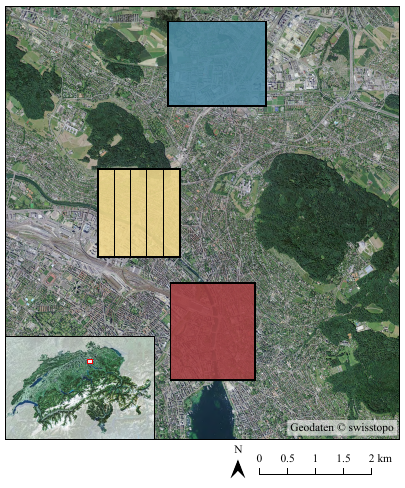}
\caption{Location of the study areas. We define three areas \roiOne (yellow), \roiTwo (red), and \roiThree (blue), where \roiOne is split into separate stripes for training, validation, and testing. We use \roiOne to perform the experiments in the main paper and \roiTwo and \roiThree to test geographical generalisation within Zurich.}
\label{fig:areas}
\end{figure}

\mypara{Motivation}
In our main experiments, we used training, validation, and test regions that are mutually exclusive but lie adjacent to each other, indicated as yellow stripes in Fig.~\ref{fig:areas}. In practice, however, it is likely that the training and test regions for 3D reconstruction will lie farther apart, be it in distinct neighborhoods of the same city or even in different cities. \Ie, their geographical context and urban design will be dissimilar, and one would expect the performance of ResDepth to deteriorate in response to the wider domain gap between training and test data, due to variations in urban topography, layout, and architecture.

\mypara{Experimental Setup}
We want to quantify how well ResDepth generalises across different districts of Zurich, without introducing any explicit measures to ensure geographic generalisation or additional training data. To that end, we define three areas, denoted as \roiOne, \roiTwo, and \roiThree, which are shown in Fig.~\ref{fig:areas}. \roiOne is equivalent to the study area used in our main experiments and is divided into separate stripes for training, validation, and testing. \roiTwo is located in the heart of Zurich and comprises the historic city centre and a stretch of the Limmat river. \roiThree consists of low residential districts, moderately high buildings, and industrial areas. \roiTwo and \roiThree cover each $\approx$2.5{$\,$km\textsuperscript{2}}.

We train ResDepth-\textit{stereo} using the three training stripes of \roiOne, following the same training procedure and settings as in the main paper. We then employ the learned model, without further fine-tuning, to refine the initial DEMs of \roiTwo and \roiThree. In addition, we simulate the situation where the training and test regions are captured by different stereo pairs. As depicted in Fig.~\ref{fig:stereopairs}, we split the set of available stereo pairs into two mutually exclusive groups $A$ and $B$, such that each group contains pairs with predominantly north-south as well as west-east oriented baselines. We use the stereo pairs in $A$ for training and the ones in $B$ for testing (and vice versa).\footnote{In particular, we run the refinement once with every stereo pair of the test set and average the statistics over all predictions.} As the model has never seen the test region and no images are shared between the training and test sets, this scenario is representative of whether the model learns meaningful features of urban design and street layout that remain valid for different geographic locations within Zurich.

\begin{table*}[t]
    \centering
    \setlength\dashlinedash{2pt}
    \setlength\dashlinegap{1.5pt}
    \setlength\arrayrulewidth{0.3pt}
    \begin{adjustbox}{max width=\textwidth}
        \begin{tabular}{@{}llccccccccc@{}}
			\toprule
			Area & Reconstruction & \multicolumn{3}{c}{Overall} & \multicolumn{3}{c}{Building pixels} & \multicolumn{3}{c}{Terrain pixels w/o water}\\
			\cmidrule(lr){3-5}\cmidrule(lr){6-8}\cmidrule(l{\tabcolsep}){9-11}
			& & MAE & RMSE & MedAE &  MAE & RMSE & MedAE & MAE & RMSE & MedAE \\
			\midrule
			\multirow{3}{*}{\roiTwo} & Initial & 4.31 & 6.18 & 2.40 & 2.80 & 4.38 & 1.67 & 5.43 & 7.23 & 3.61 \\
			& ResDepth-\textit{stereo} ($A$$\;\rightarrow\;$train, $B$$\;\rightarrow\;$test)          & 3.05 & 4.91 & 1.49 & 2.56 & 4.50 & 1.15 & 3.28 & 5.07 & 1.72 \\
			& ResDepth-\textit{stereo} ($B$$\;\rightarrow\;$train, $A$$\;\rightarrow\;$test)          & 3.00 & 4.90 & 1.43 & 2.57 & 4.51 & 1.14 & 3.23 & 5.09 & 1.63 \\
			[0.7ex]
            \hdashline\noalign{\vskip 0.7ex}
			\multirow{3}{*}{\roiThree} & Initial  & 2.43 & 4.11 & 1.14 & 2.17 & 3.77 & 0.94 & 2.53 & 4.23 & 1.23 \\
			& ResDepth-\textit{stereo} ($A$$\;\rightarrow\;$train, $B$$\;\rightarrow\;$test)          & 2.03 & 3.73 & 0.92 & 2.59 & 4.46 & 1.11 & 1.81 & 3.41 & 0.86 \\
			& ResDepth-\textit{stereo} ($B$$\;\rightarrow\;$train, $A$$\;\rightarrow\;$test)          & 2.02 & 3.74 & 0.89 & 2.52 & 4.37 & 1.08 & 1.82 & 3.46 & 0.83 \\
		    \bottomrule
	    \end{tabular}
    \end{adjustbox}
    \vspace*{\floatsep}
    \caption{Quantitative DEM refinement results when generalising across space within Zurich (in [m]). We evaluate the mean absolute error (MAE), the root mean square error (RMSE), and the median absolute error (MedAE) over all pixels, building pixels only, and terrain pixels excluding water bodies. The image sets $A$ and $B$ refer to Fig.~\ref{fig:stereopairs}.\vspace{1cm}}
    \label{tab:generalisation}
\end{table*}

\begin{figure*}[ht]
    \subfloat[Example panchromatic image]{
    \includegraphics[width=\columnwidth]{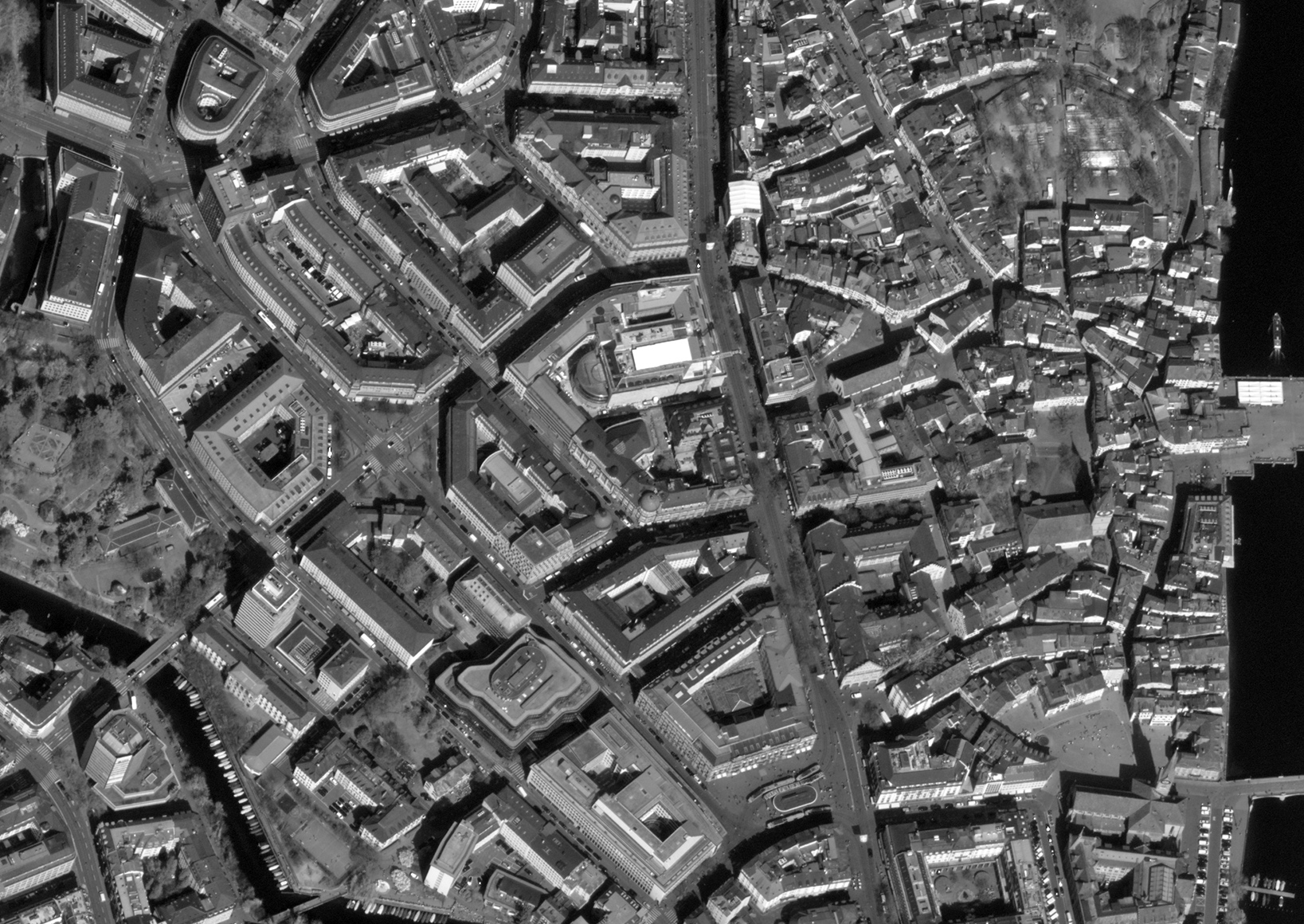}}
    \hfill
    \subfloat[Initial DEM]{
    \includegraphics[width=\columnwidth]{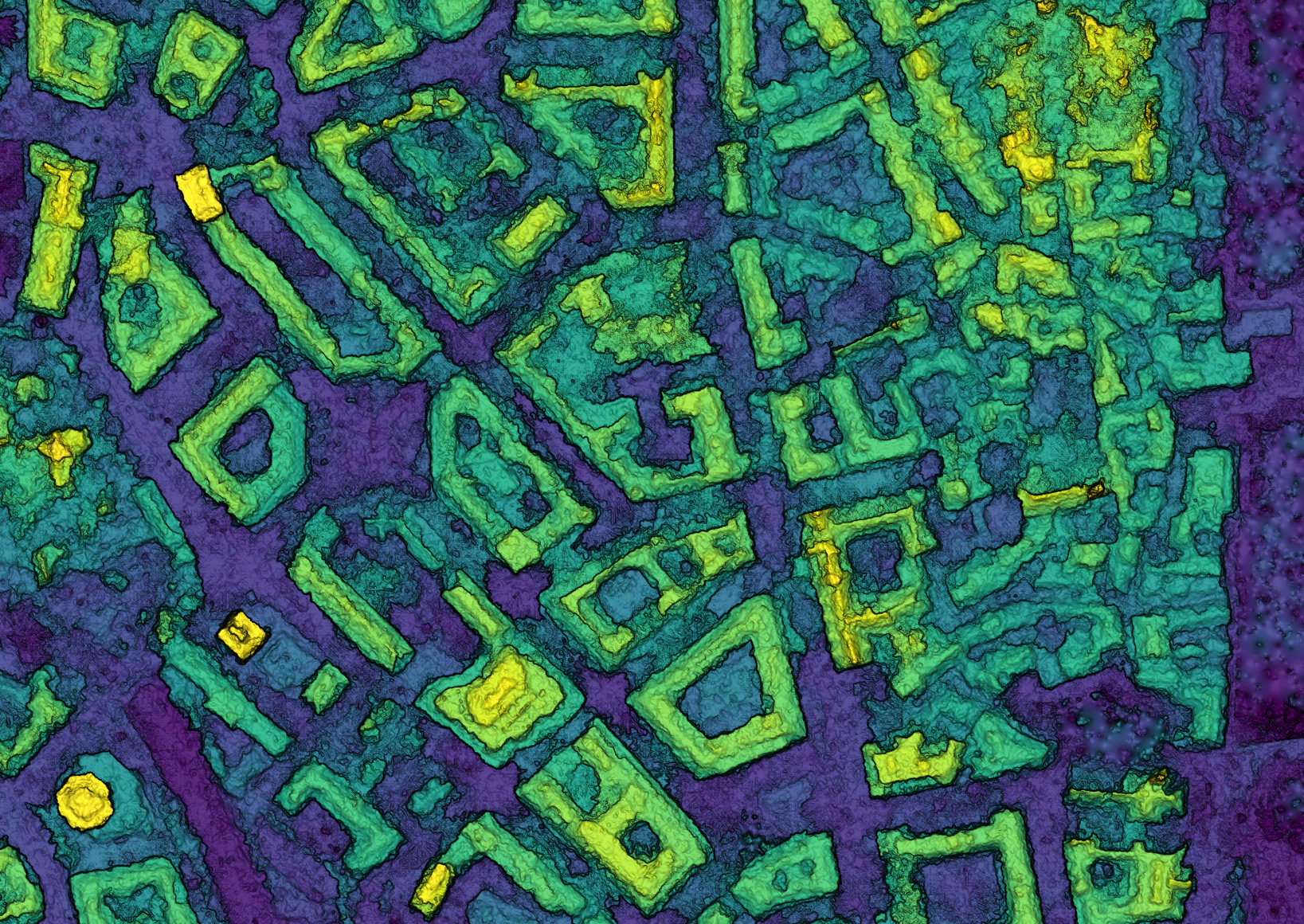}}
    
    \subfloat[Refined DEM]{
    \includegraphics[width=\columnwidth]{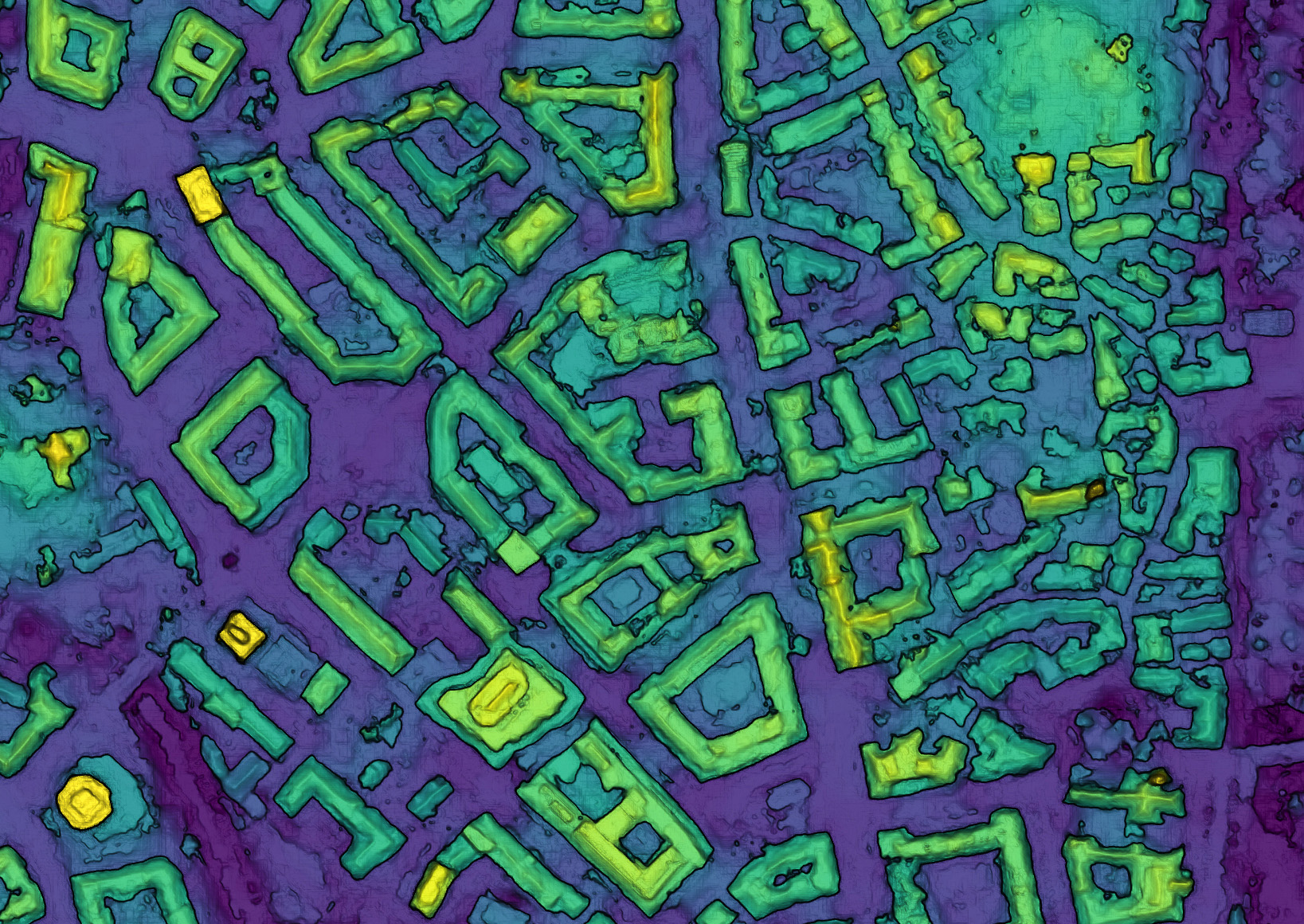}}
    \hfill
    \subfloat[Ground truth DEM]{
    \includegraphics[width=\columnwidth]{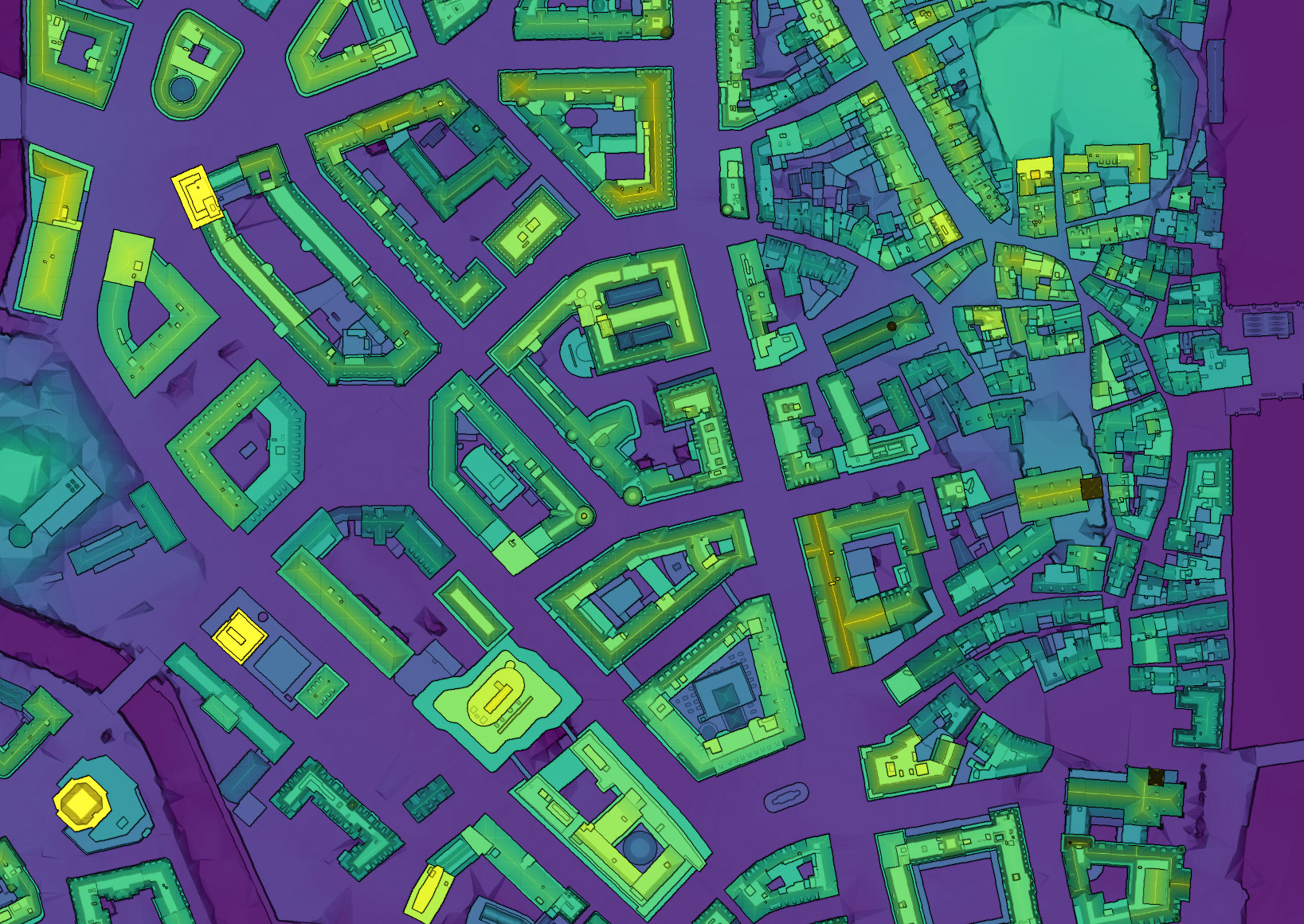}}
    \vspace*{5mm}
    \caption{Geographical generalisation between different regions of Zurich. Heights are color-coded from blue to green to yellow. We show results for the city centre of Zurich, located in \roiTwo. See Fig~\ref{fig:zurich_inner_city_3d} for the corresponding 3D view.}
    \label{fig:zurich_inner_city}
\end{figure*}

\begin{figure*}[ht]
    \centering
    \subfloat[Initial DEM]{
    \includegraphics[trim=0 20 0 30,clip, scale=0.84]{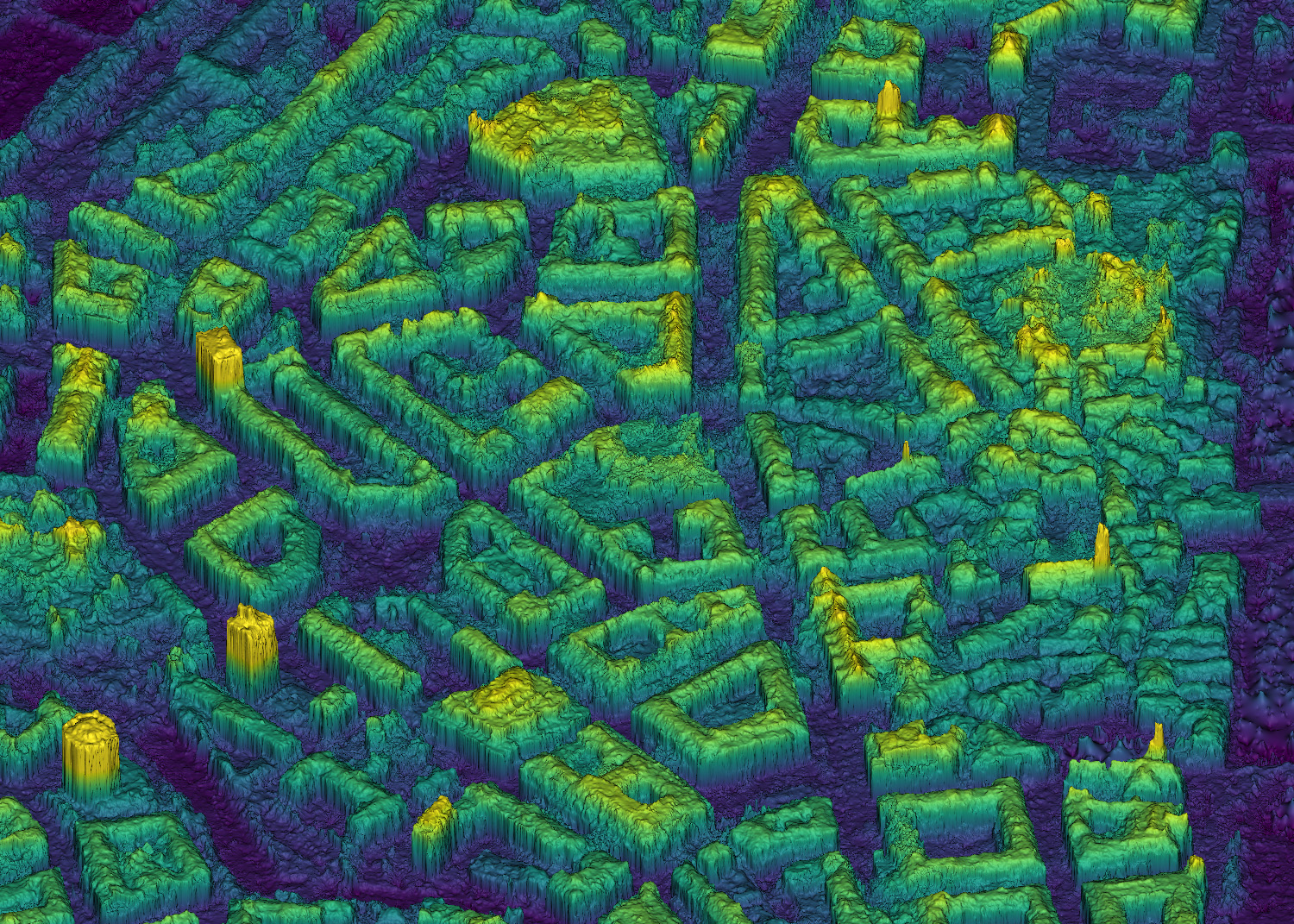}}
    
    \subfloat[Refined DEM]{
    \includegraphics[trim=0 20 0 30,clip, scale=0.84]{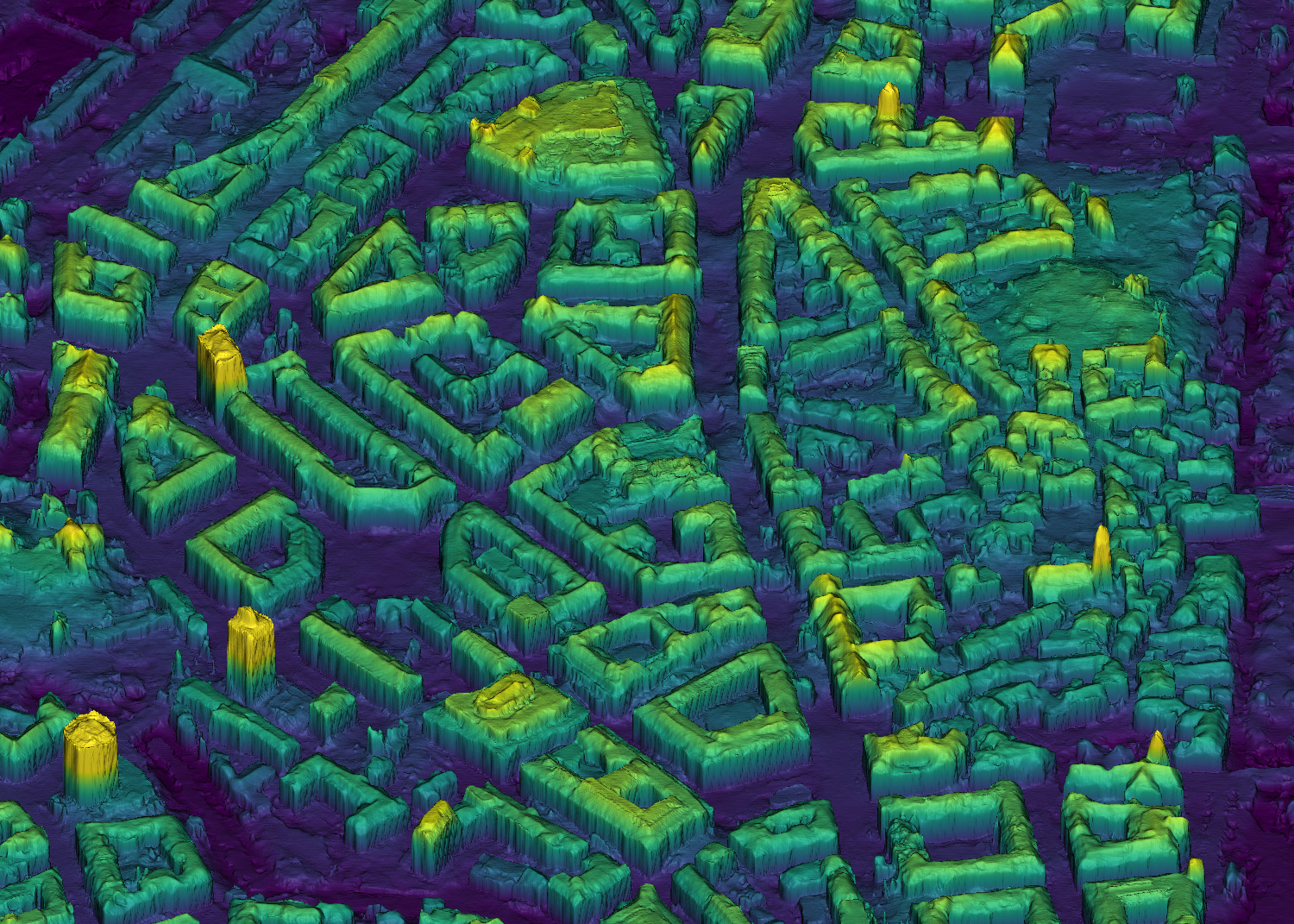}}
    
    \subfloat[Ground truth DEM]{
    \includegraphics[trim=0 20 0 30,clip, scale=0.84]{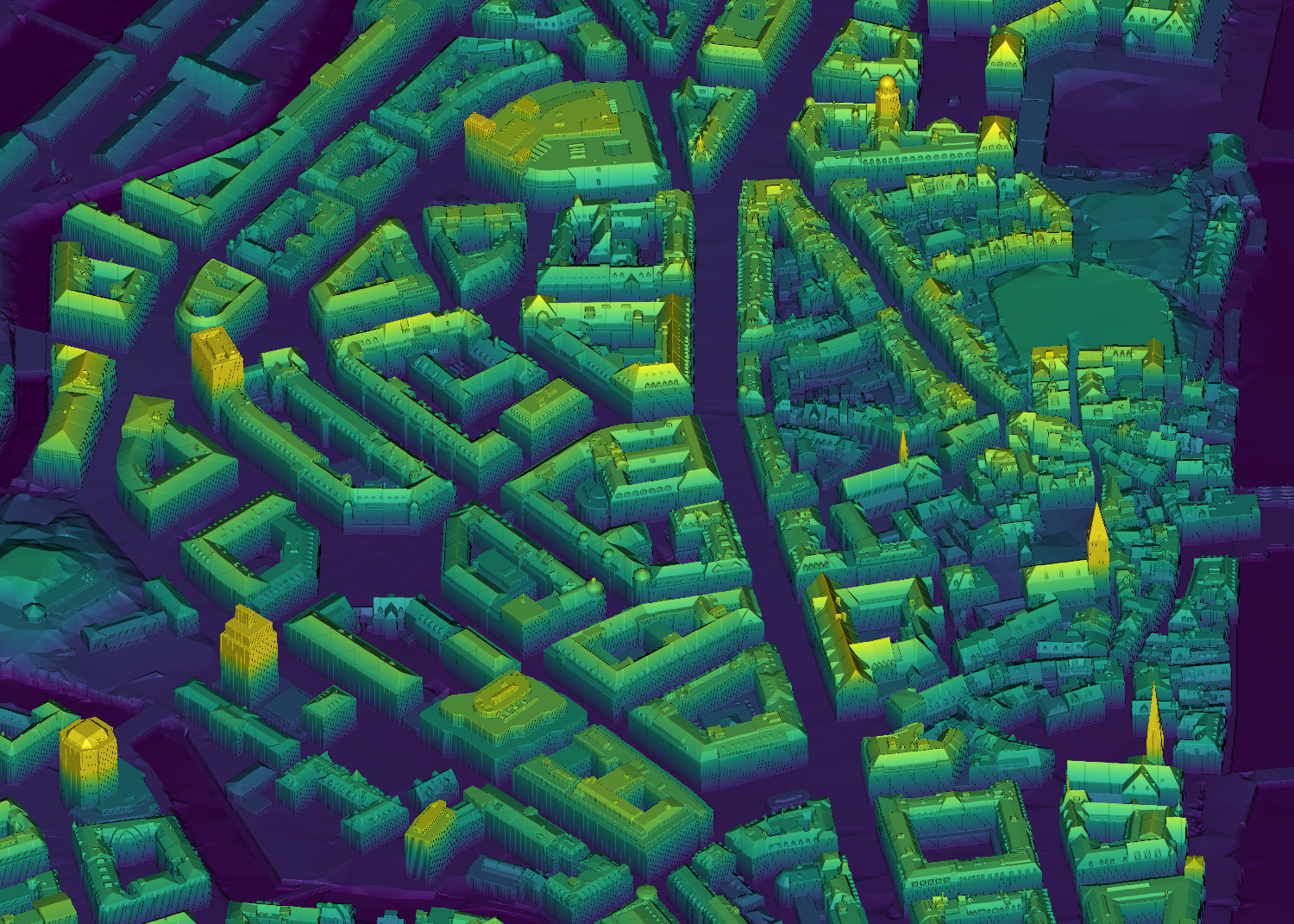}}
    \vspace*{5mm}
    \caption{Geographical generalisation between different regions in Zurich. Heights are color-coded from blue to green to yellow. The picture shows a 3D view of the refined scene depicted in Figure~\ref{fig:zurich_inner_city}.}
    \label{fig:zurich_inner_city_3d}
\end{figure*}

\begin{figure*}[hp]
    \centering
    \subfloat[Example panchromatic image]{
    \includegraphics[scale=0.97]{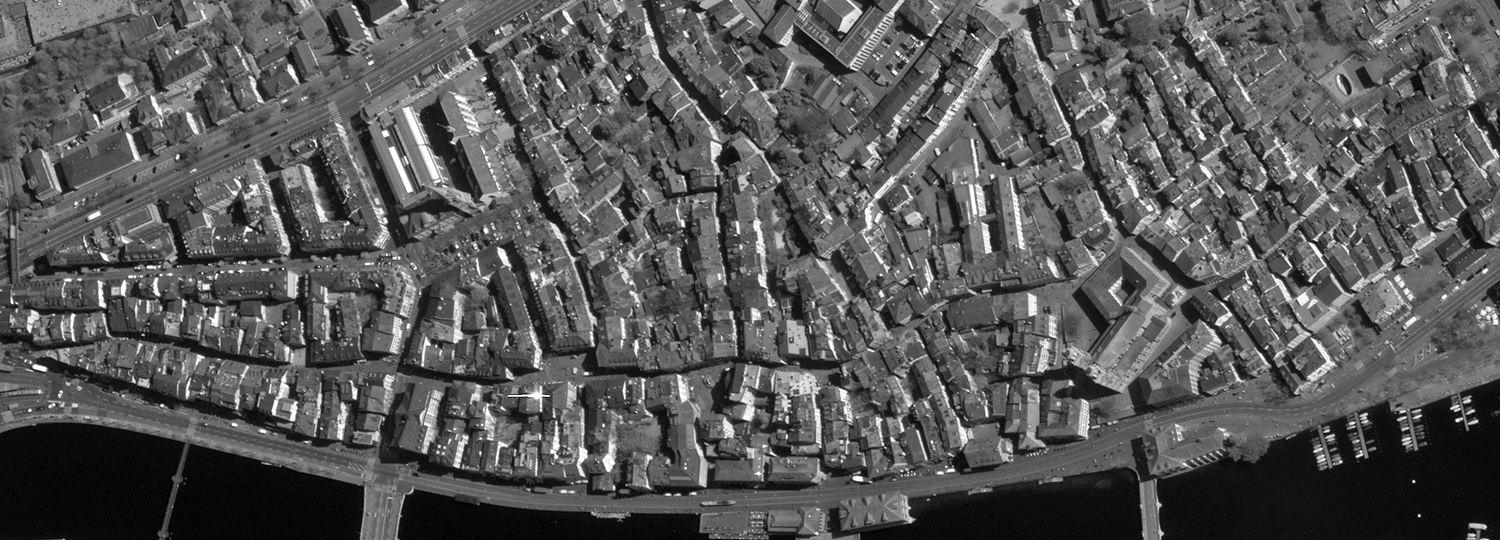}}
    
    \subfloat[Initial DEM]{
    \includegraphics[scale=0.97]{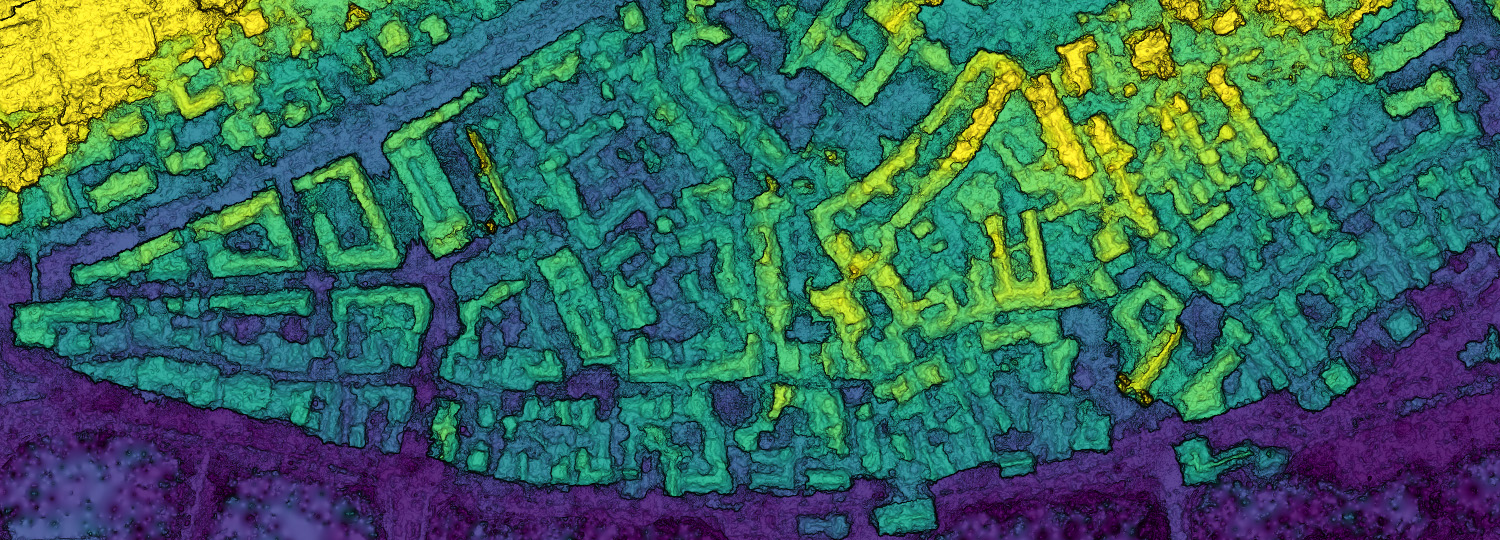}}
    
    \subfloat[Refined DEM]{
    \includegraphics[scale=0.97]{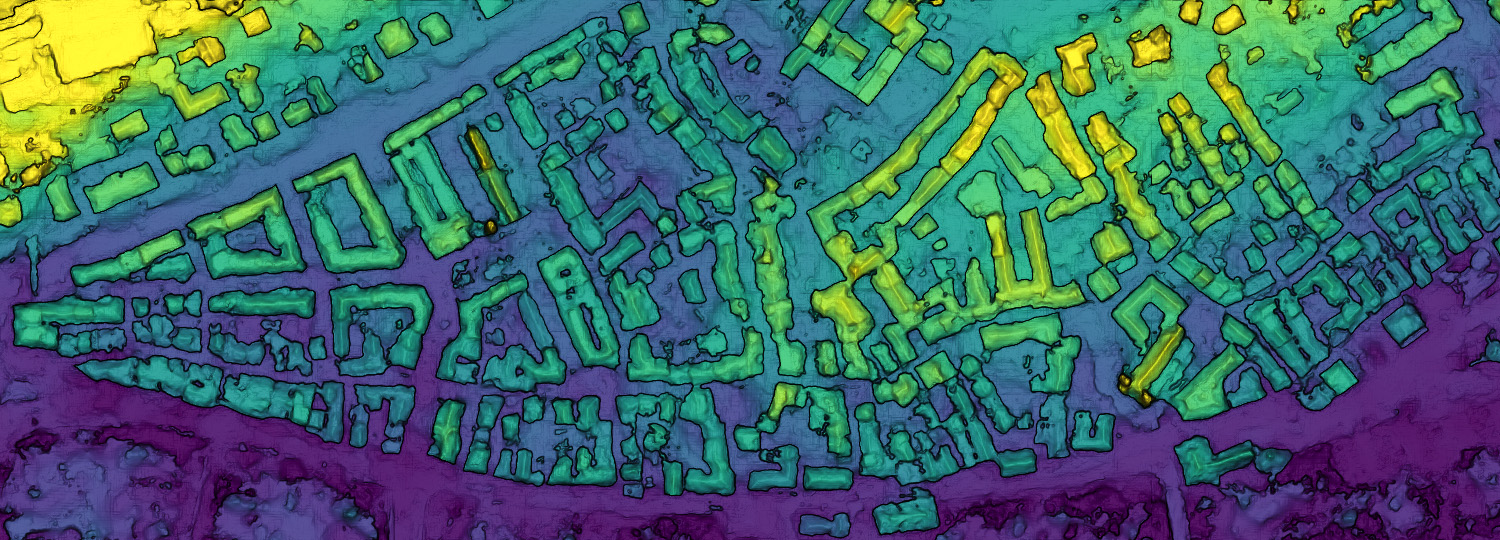}}
    
    \subfloat[Ground truth DEM]{
    \includegraphics[scale=0.97]{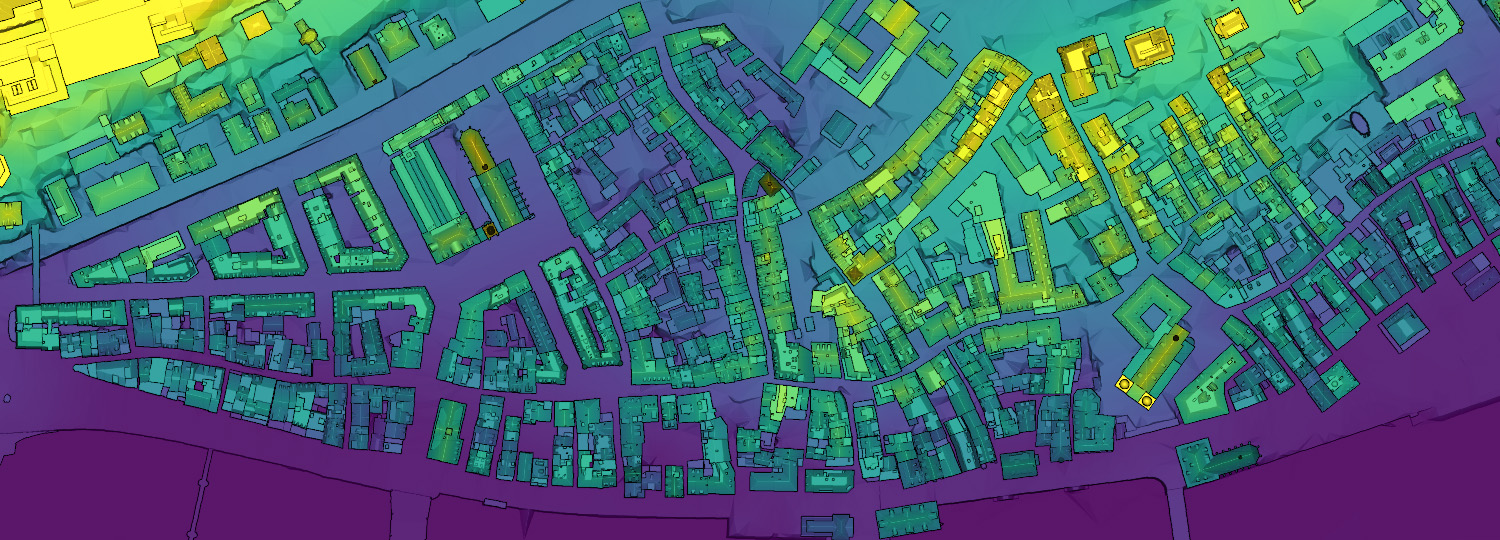}}
    \vspace*{5mm}
    \caption{Geographical generalisation between different areas in Zurich. Heights are color-coded from blue to green to yellow. Refinement results for the historic city centre of Zurich located in \roiTwo.}
    \label{fig:zurich_niederdorf}
\end{figure*}

\begin{figure*}[hp]
    \centering
    \subfloat[Example panchromatic image]{
    \includegraphics[width=0.85\columnwidth]{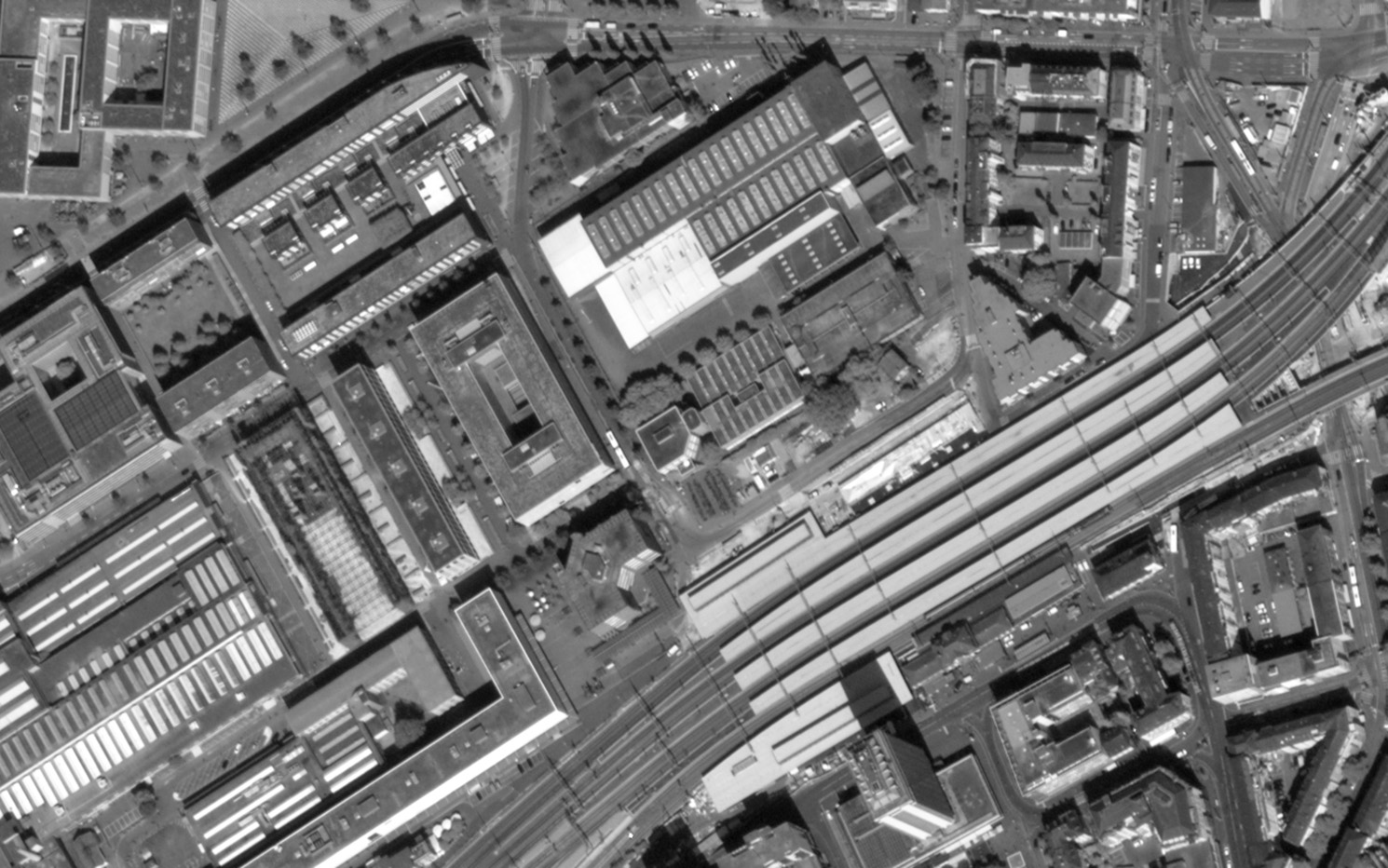}
    \quad
    \includegraphics[width=0.85\columnwidth]{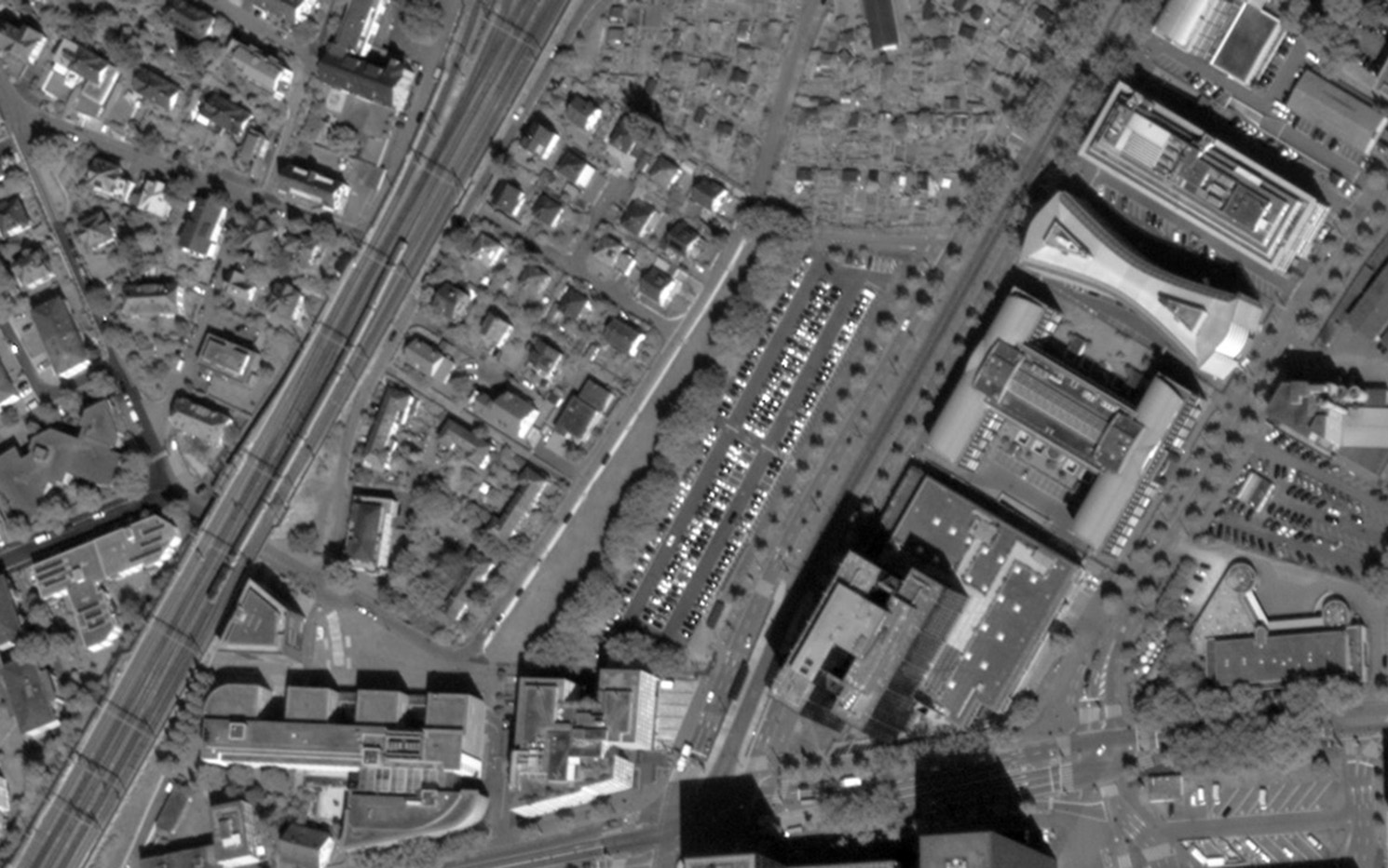}
    }
    \\
    \subfloat[Initial DEM]{
    \includegraphics[width=0.85\columnwidth]{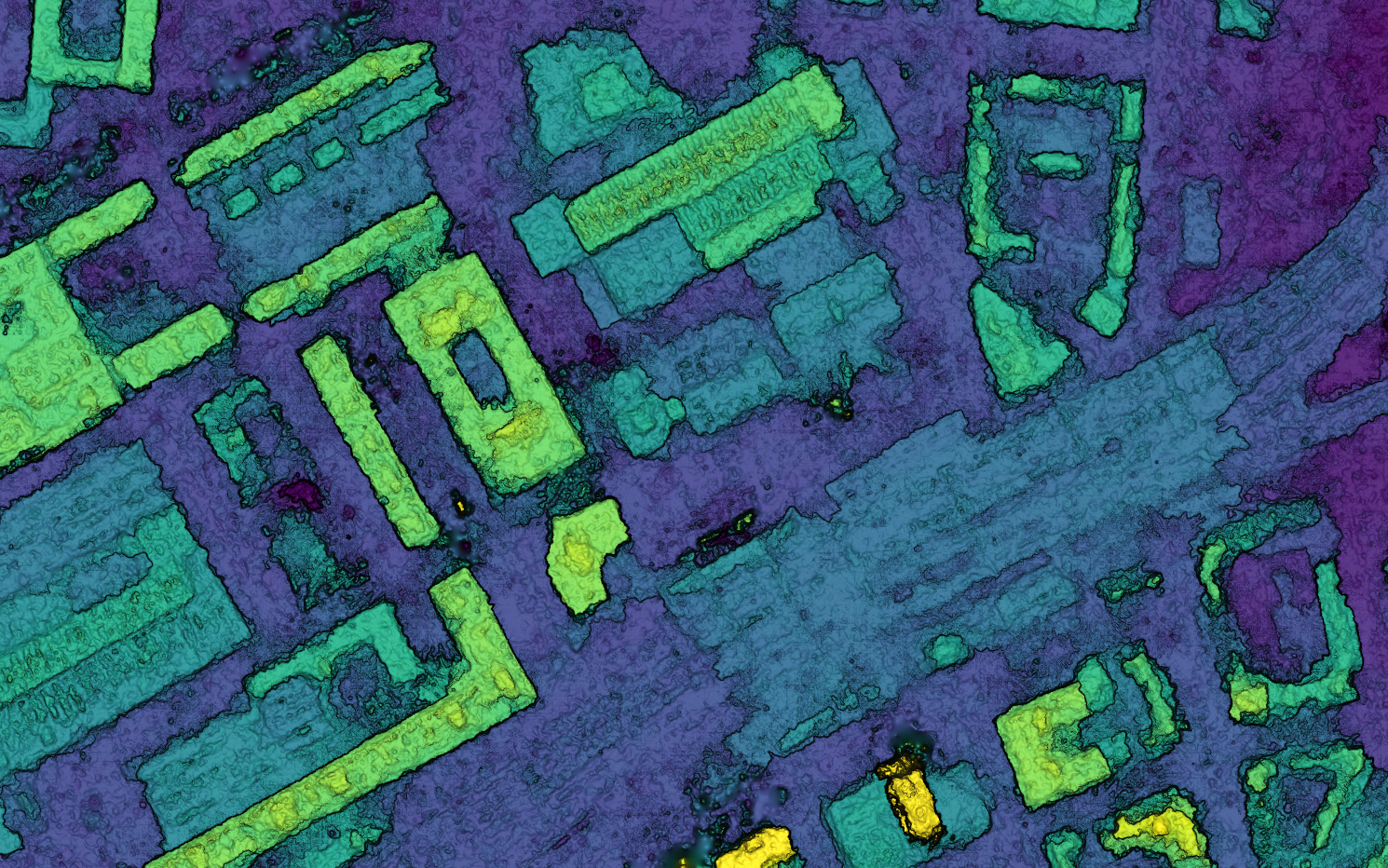}
    \quad
    \includegraphics[width=0.85\columnwidth]{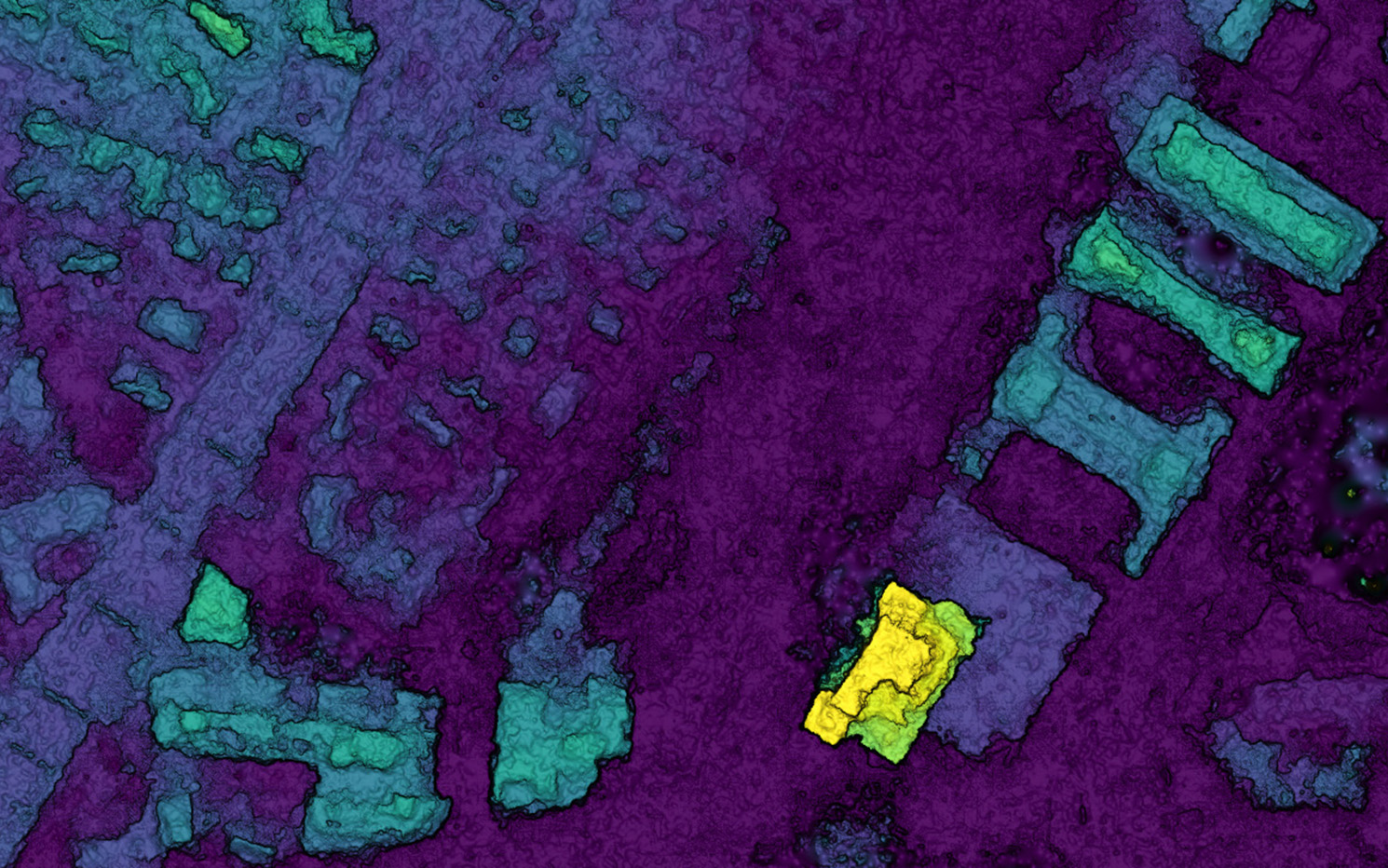}
    }
    \\
    \subfloat[Refined DEM]{
    \includegraphics[width=0.85\columnwidth]{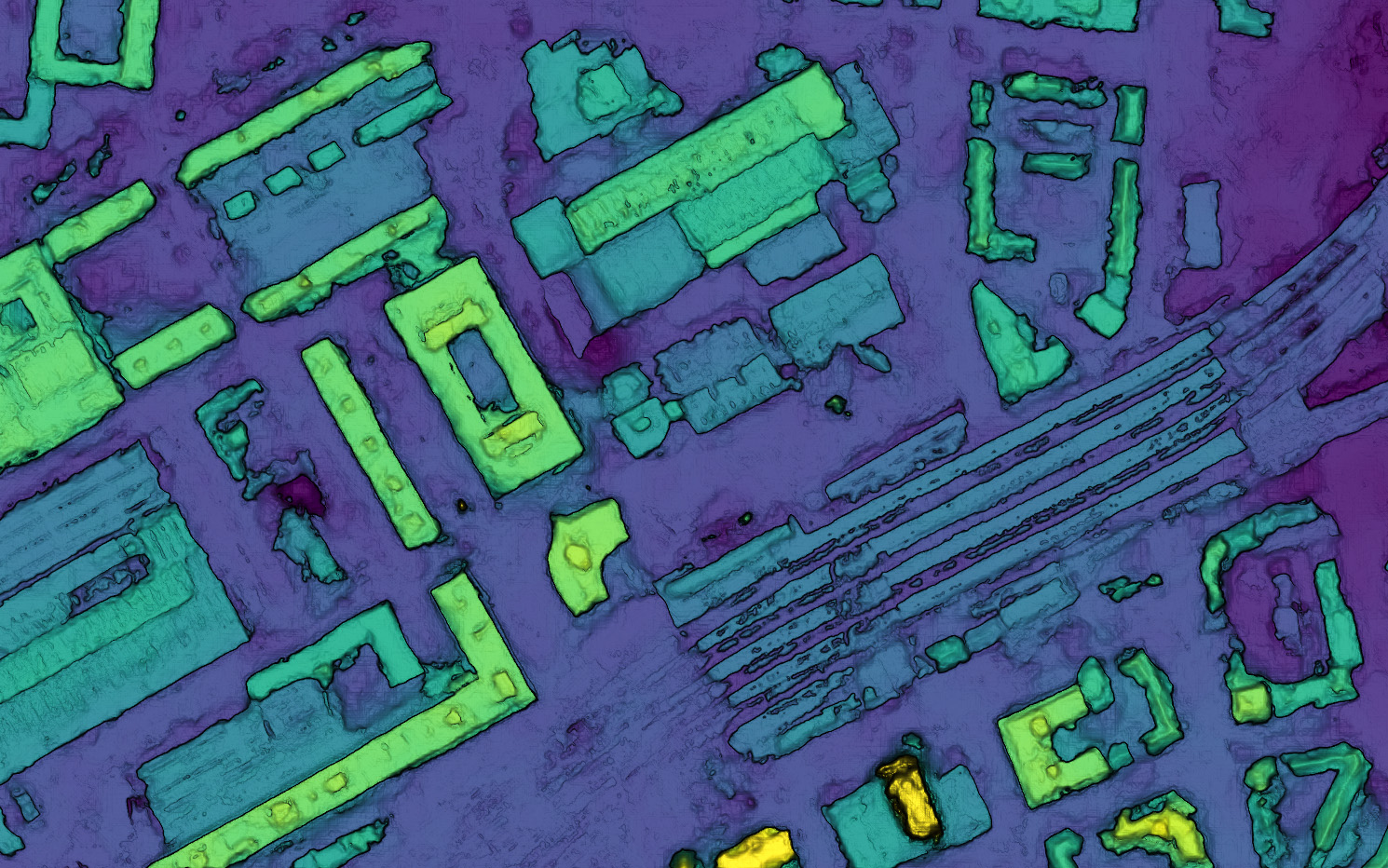}
    \quad
    \includegraphics[width=0.85\columnwidth]{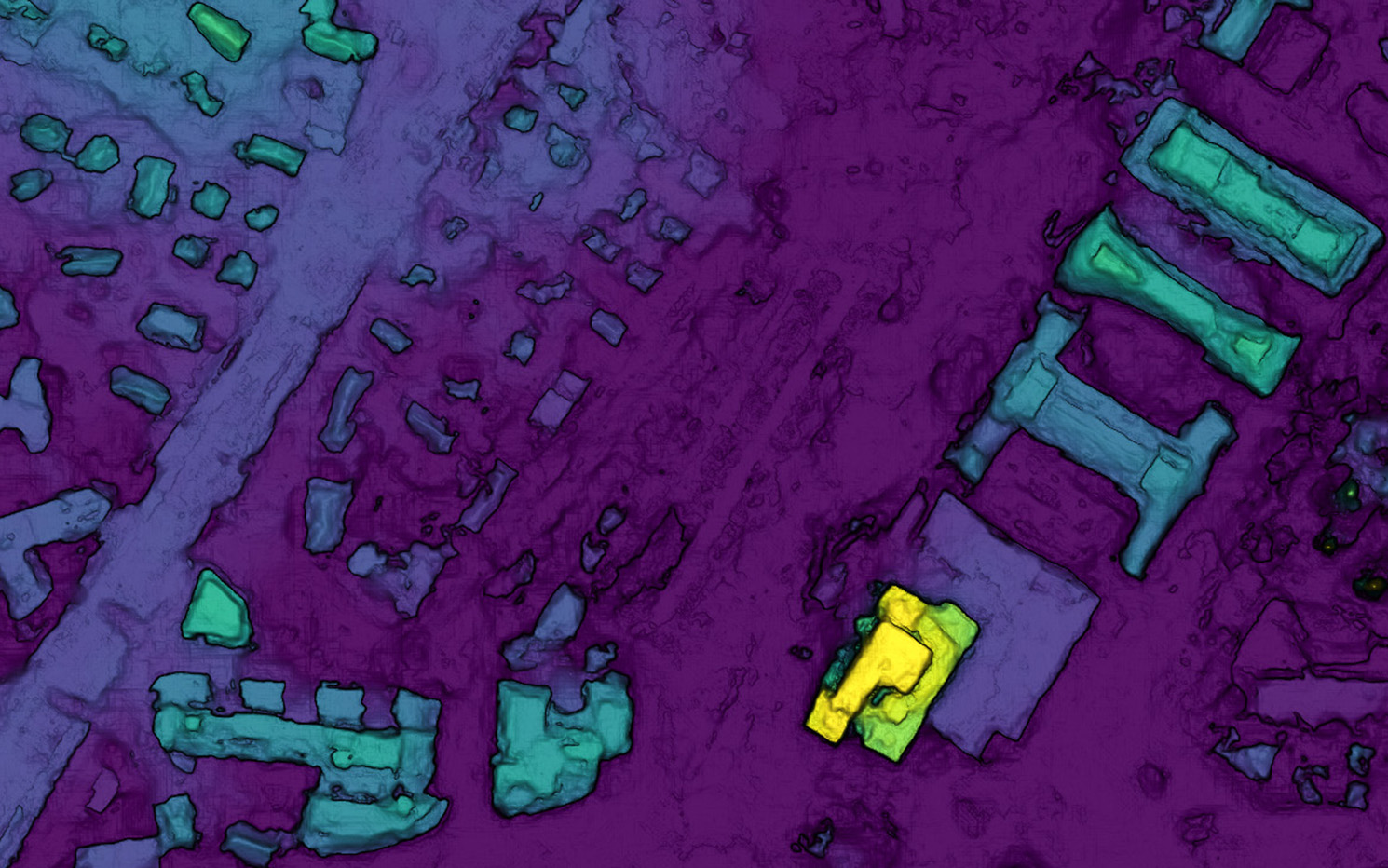}
    }
    \\
    \subfloat[Ground truth DEM]{
    \includegraphics[width=0.85\columnwidth]{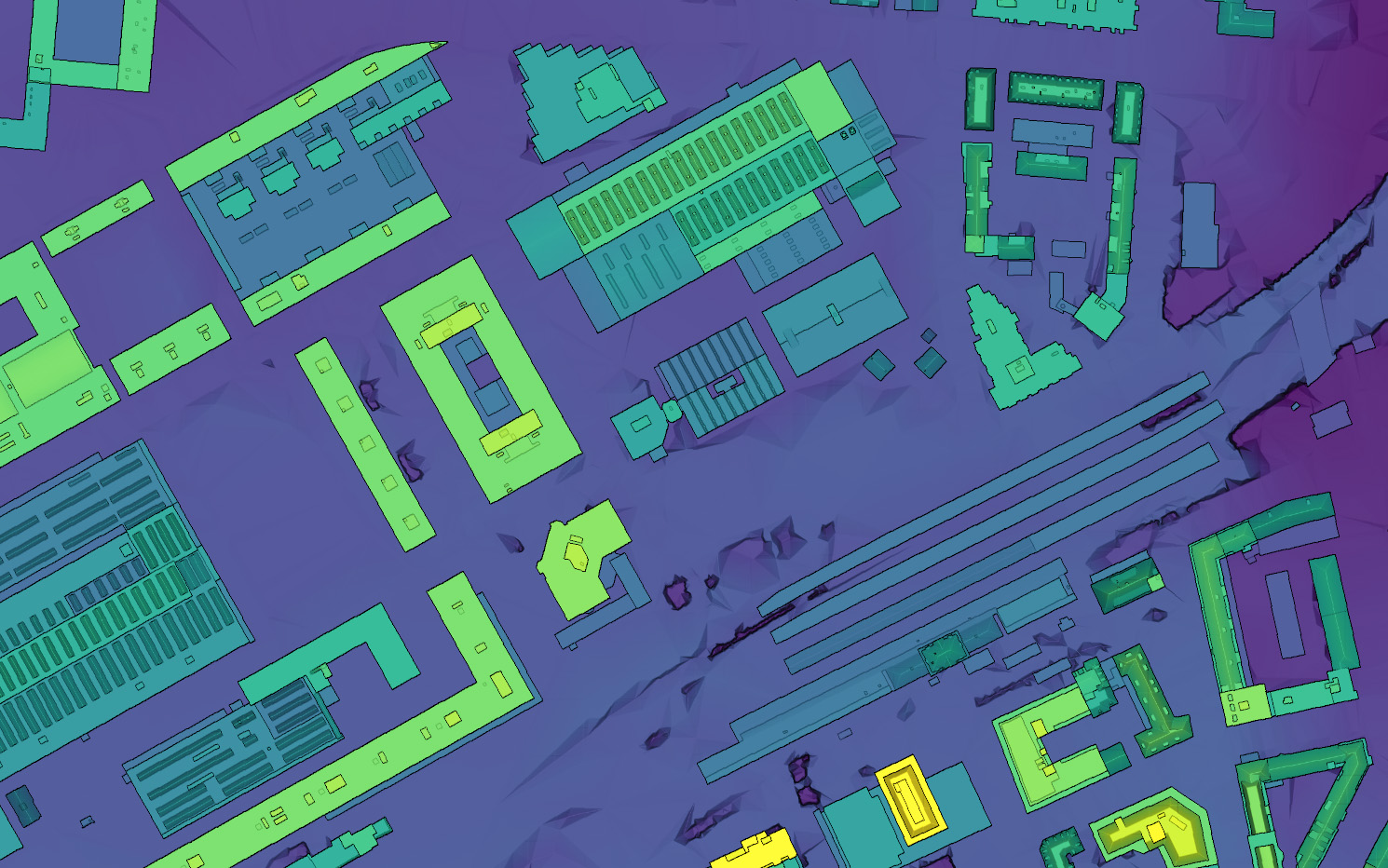}
    \quad
    \includegraphics[width=0.85\columnwidth]{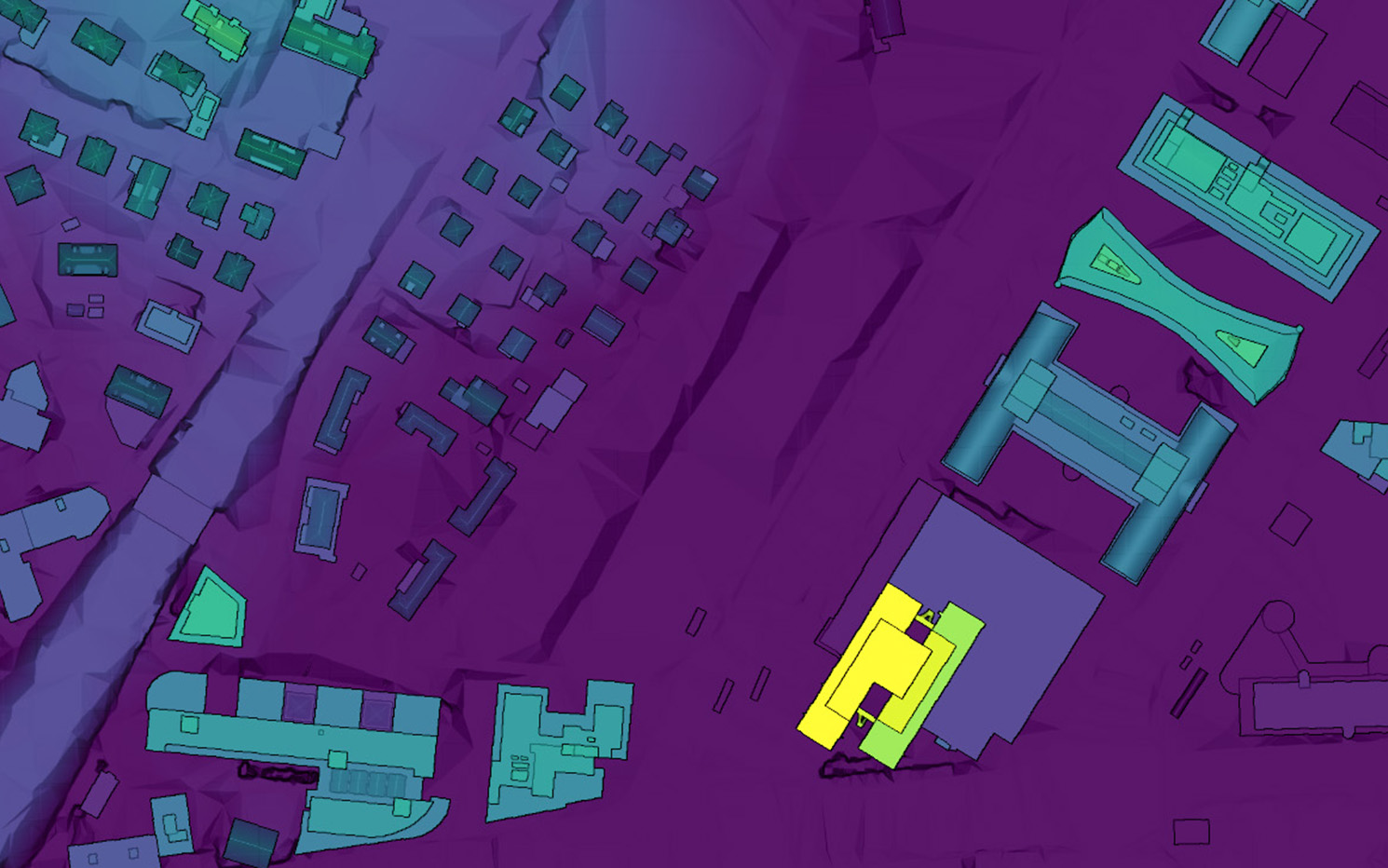}
    }
    \vspace*{5mm}
    \caption{Geographical generalisation between different regions in Zurich. Heights are color-coded from blue to green to yellow. Refinement results for two scenes in \roiThree.}
    \label{fig:oerlikon}
\end{figure*}

\mypara{Results}
We report quantitative results in Table~\ref{tab:generalisation} and show visual examples in
Fig.~\ref{fig:zurich_inner_city} to Fig.~\ref{fig:oerlikon}. ResDepth-\textit{stereo} improves the reconstruction quality in \roiTwo and \roiThree both quantitatively and qualitatively. Compared to the initial reconstruction, it reduces the overall MAE by 30\% and 16\%, irrespective of which image sets were used during training and inference. Visually, the learned model performs reasonably well when generalising over larger distances. Even for challenging scenes, such as the intricate and closely spaced buildings of the city centre of Zurich (see Fig.~\ref{fig:zurich_inner_city} to Fig.~\ref{fig:zurich_niederdorf}), the model is able to sharpen building outlines and separate built-up areas into plausible individual units. Strikingly, the model also recovers surface details missed in the initial reconstruction, such as the roofs of the railway platform in \roiThree (see Fig.~\ref{fig:oerlikon}, \nth{1} column). Although the model has never seen a railway station during training, it correctly reconstructs height discontinuities between the roofs and rails and aligns them with the image gradients in the satellite views.

Recall that we have applied the same ResDepth-\textit{stereo} model to the test stripe of \roiOne, see Table~\ref{tab:ablation} (last two rows). For that test region, the model reduces the overall MAE by more than 50\%. The considerably higher gain in accuracy for the test stripe in \roiOne compared to \roiTwo and \roiThree is due to the data-driven nature of machine learning: since the training data consists of DEM samples that are geographically very close to the test region in \roiOne, the training data distribution better reflects the urban characteristics and peculiarities of \roiOne than of \roiTwo and \roiThree. The domain gap between training and test sites is particularly pronounced for building pixels. While the model yields quantitative gains by $\approx$20\% for building pixels in the test stripe of \roiOne, it results in only marginally lower error metrics in \roiTwo w.r.t.\ the initial DEM. For building pixels in \roiThree, we report an increase of the MAE by almost $\approx$20\% compared to the initial DEM. Qualitatively, we observe that ResDepth-\textit{stereo} has difficulties with low and small residential buildings and allotments. As an example, Fig.~\ref{fig:oerlikon} (\nth{2}~column) shows that ResDepth-\textit{stereo} fails to recover small buildings.

Beyond the domain gap between different districts of Zurich, we postulate that our DEM normalization scheme based on a globally computed mean height and standard deviation introduces biases that negatively affect the prediction. To rescale the data, we compute a single scale factor across all DEM patches sampled in the raining stripes of \roiOne, which exhibit height variations on the order of 100$\,$m over distances \textless1$\,$km. At test time, we globally center the initial DEM to its mean height in order to minimize the effect of absolute vertical positioning. Then, we rescale the centered heights with the standard deviation estimated during training. However, the hilly topography of \roiOne results in an excessively large scale factor for the rather flat terrain of the test region \roiThree---after rescaling, a family home in \roiThree is much lower than a family home with the same metric height in \roiOne. Therefore, to preserve a (relative) notion of scale, we propose an alternative, local standardization scheme. We show in our follow-up work~\citeSupp{stucker2021resdepth} that this new strategy is indeed adequate to minimize the influence of terrain and yields better generalisation performance.

\subsection{Impact of the Preceding Stereo Matching Method}

\mypara{Motivation}
Our ResDepth network is generic and can thus be used in combination with any stereo matching algorithm. After concordant training, the network implicitly learns to adapt to the specific biases and error patterns induced by the preceding stereo matcher. We showcase this flexibility of our approach with two state-of-the-art stereo matching methods. First, we combine ResDepth with an adapted version of the tSGM stereo matcher~\cite{rothermel2012sure}, a hierarchical semi-global matching method originally developed for terrestrial and airborne photogrammetry. This is the matcher that has been used throughout all previous experiments. Second, we use ResDepth to complement the \textit{s2p} satellite stereo pipeline~\citeSupp{deFranchis2014automatic}.

\mypara{\textit{s2p} Processing}
To ensure that the results are as comparable as possible, we process exactly the same stereo pairs with the \textit{s2p} processing pipeline as previously processed with the tSGM matcher. Furthermore, we provide the same bias-corrected rational polynomial coefficients (RPC) of the camera projection model.\footnote{Note that, we manually enable/disable the local pointing correction of \textit{s2p}, because we found the method to work better with this additional correction only for certain stereo pairs.} For our data, \textit{s2p} does not work "out-of-the-box" and produces reconstructions with rather poor quality. On the one hand, the individual stereo point clouds exhibit a significantly higher noise level than those derived with the tSGM matcher. On the other hand, \textit{s2p} splits each rectified stereo pair into a regular grid of image tiles and performs subsequent image matching and triangulation for each pair of tiles separately. However, \textit{s2p} often fails to correctly estimate the disparity search range. As a consequence, the triangulated image correspondences of a tile exhibit significant height bias. To mitigate those effects, we had to manually tune the disparity search range for the tiles. Despite careful manual tuning of the parameters, the resulting stereo point clouds remain rather noisy. We thus manually delete obvious outliers before merging the filtered point clouds into a single multi-view DEM. Lastly, we perform another round of cleaning by applying a median filter (with kernel size 5$\times$5 pixels) to the merged DEM.

\begin{figure*}[ht]
    \centering
    \subfloat[]{
    \includegraphics[width=0.24\textwidth]{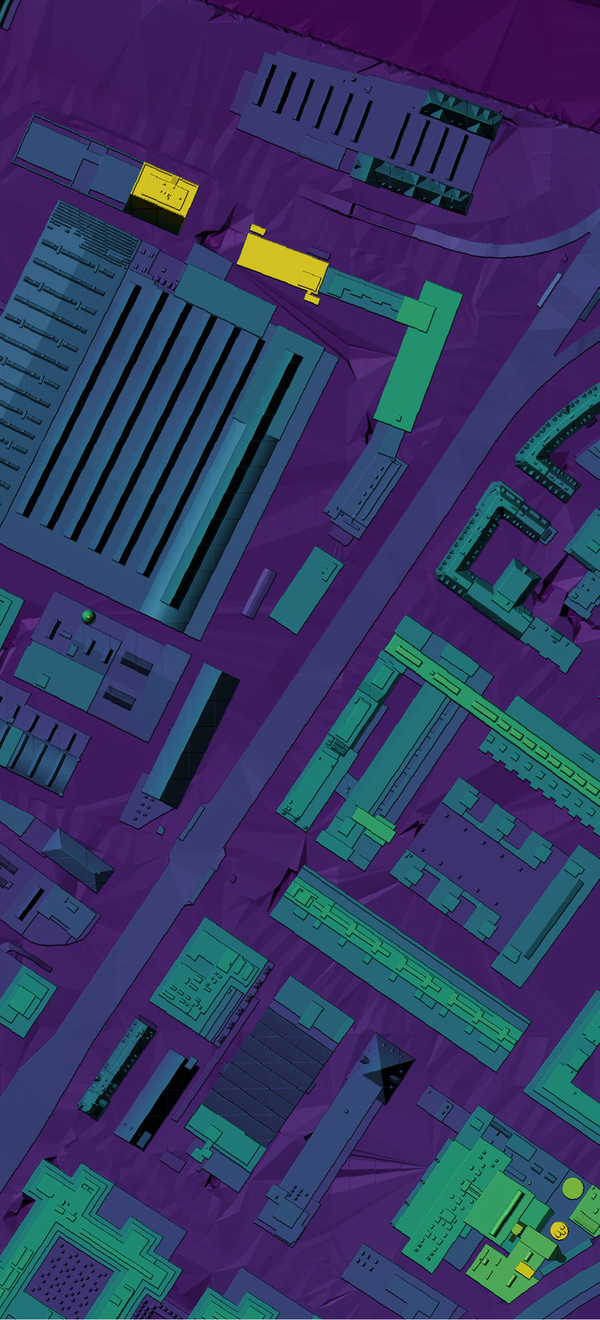}}
    \hfil
    \subfloat[]{
    \includegraphics[width=0.24\textwidth]{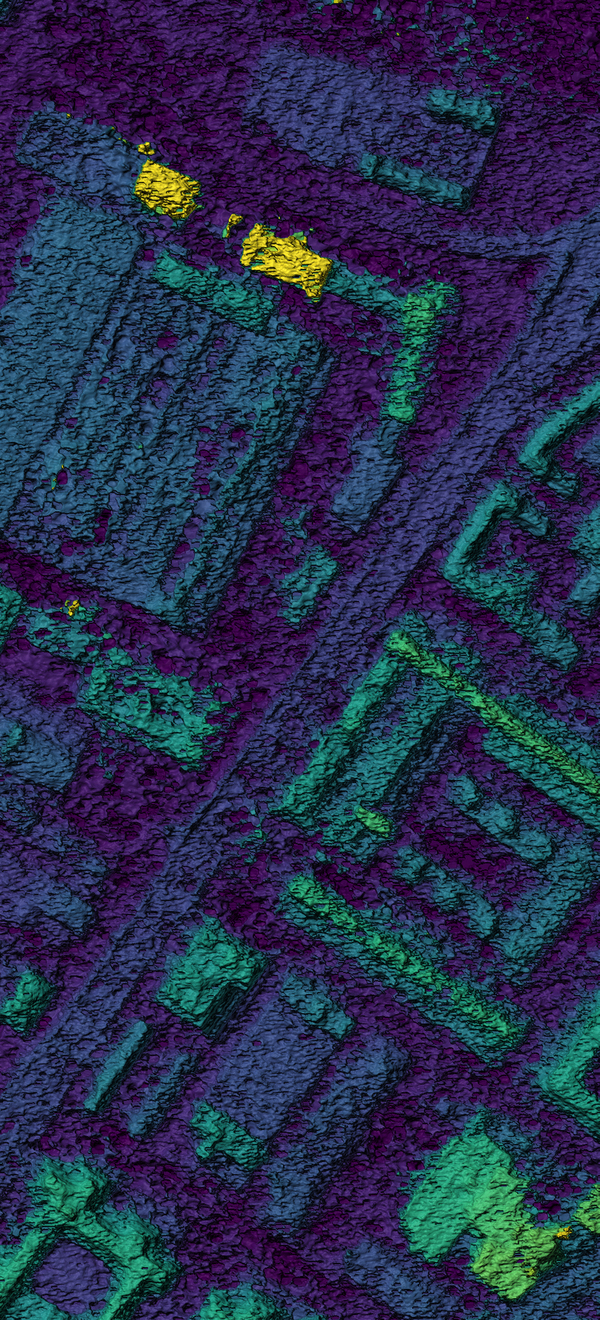}\label{fig:s2p_initial}}
    \hfil
    \subfloat[]{
    \includegraphics[width=0.24\textwidth]{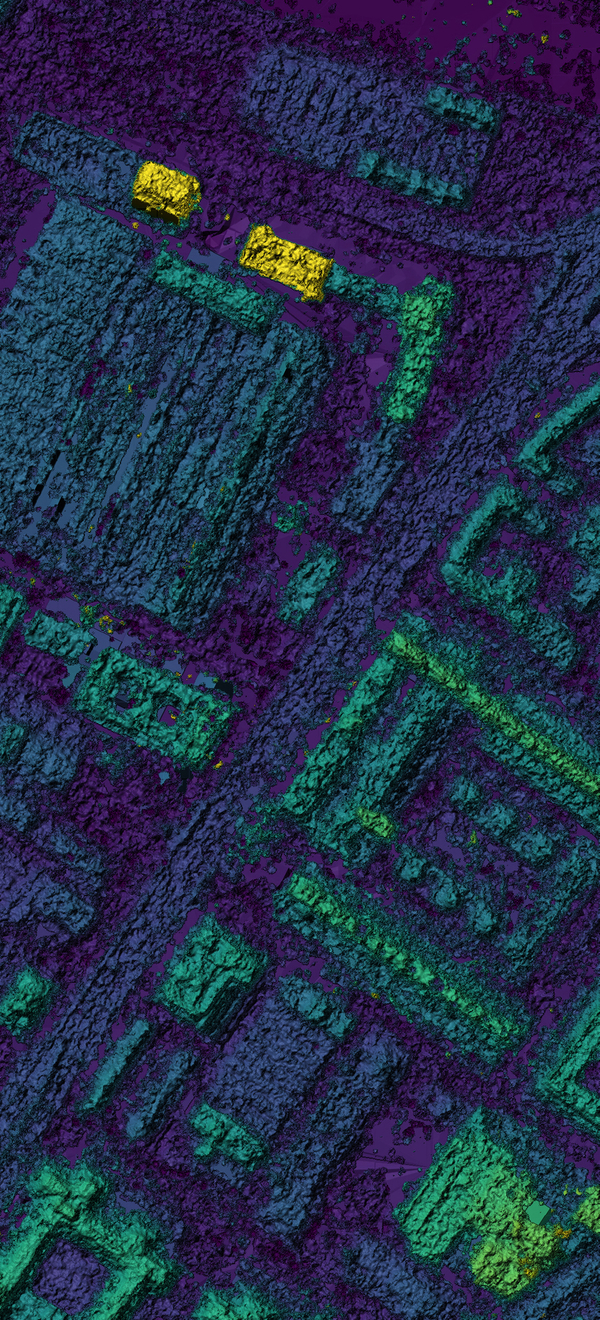}\label{fig:tSGM_initial}}
    \\
    \subfloat[]{
    \includegraphics[width=0.24\textwidth]{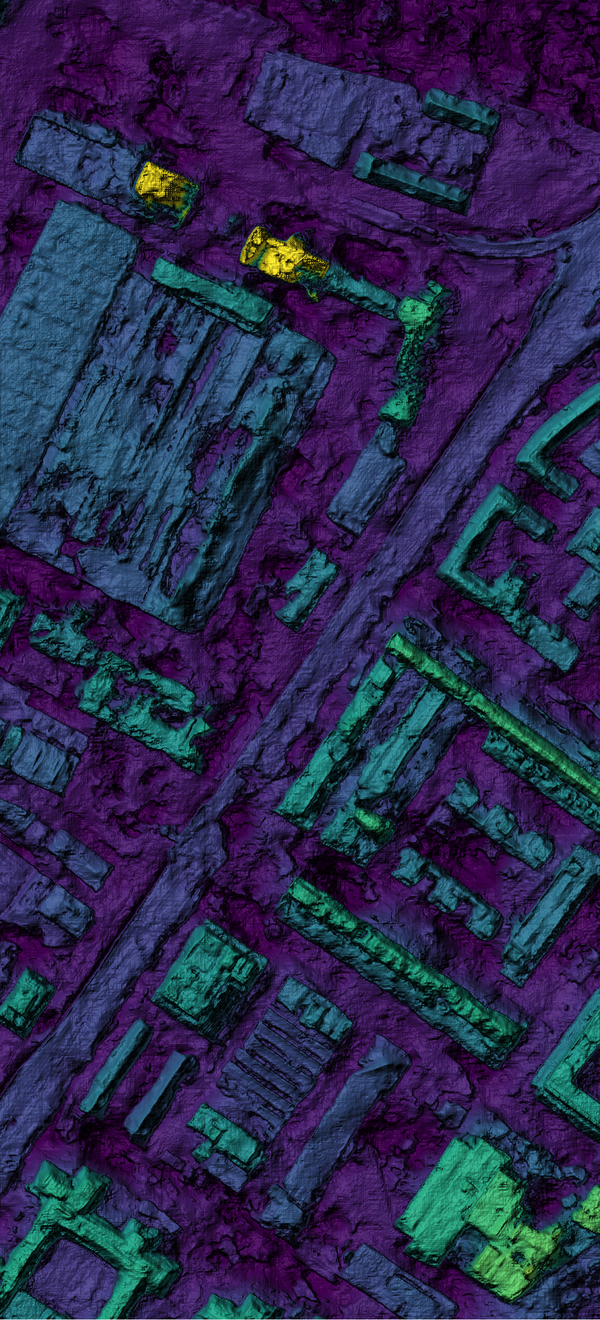}\label{fig:s2p_refined_tSGM_trained}}
    \hfil
    \subfloat[]{
    \includegraphics[width=0.24\textwidth]{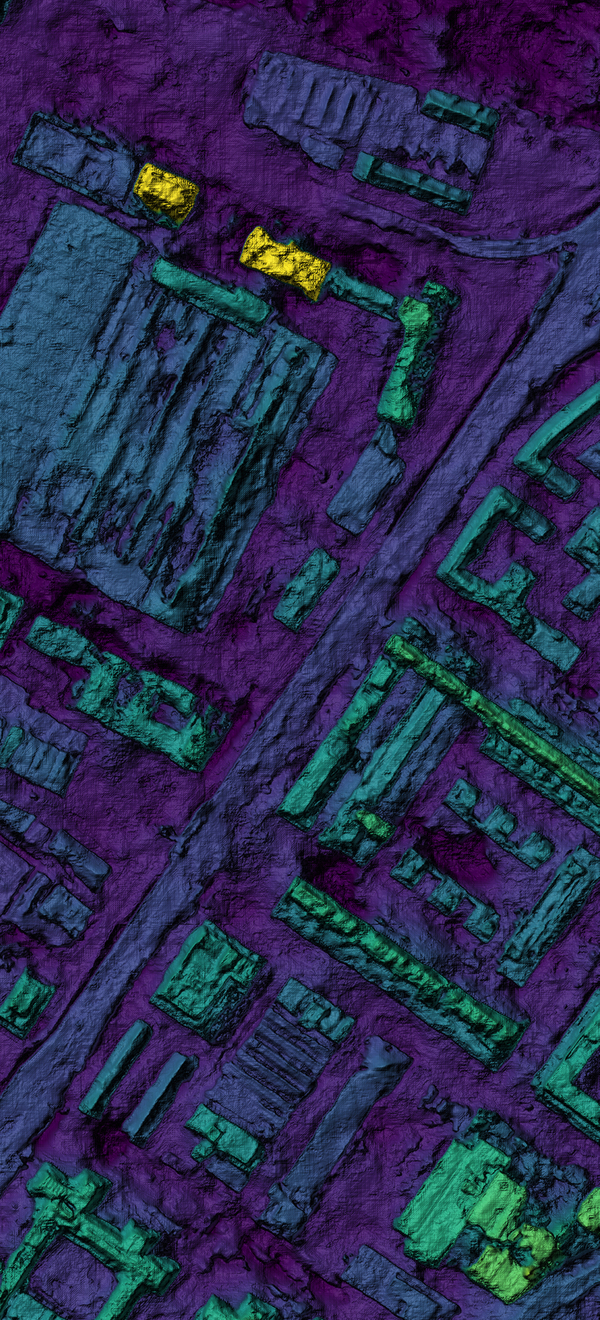}\label{fig:s2p_refined_s2p_trained}}
    \hfil
    \subfloat[]{
    \includegraphics[width=0.24\textwidth]{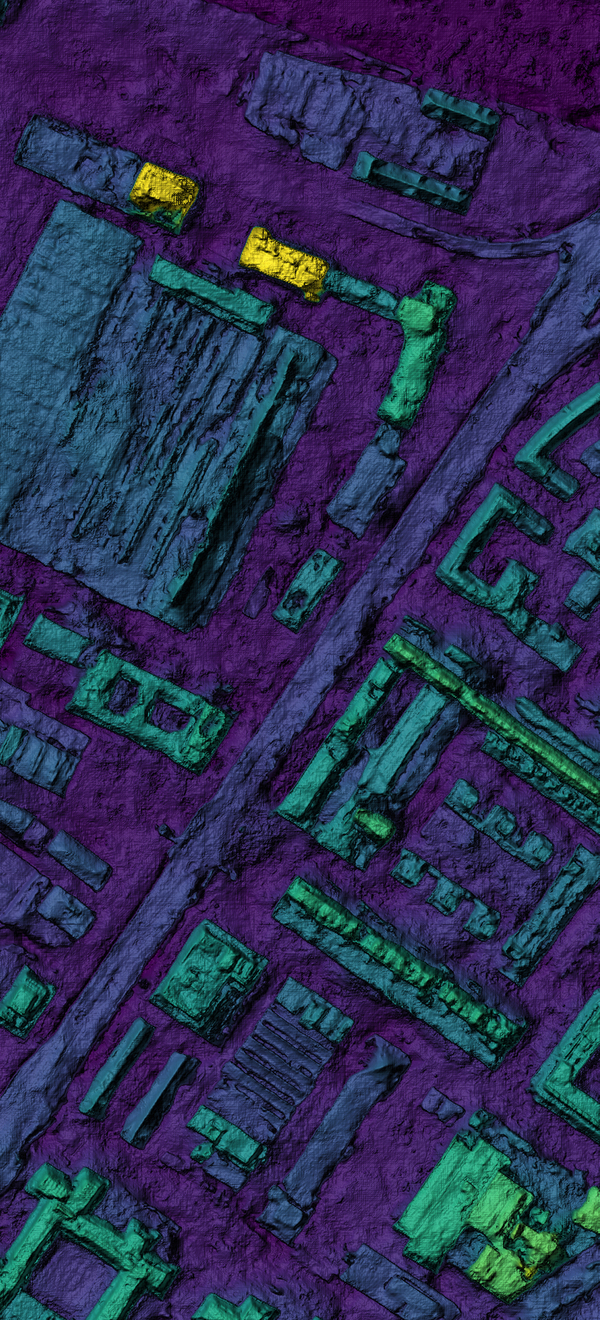}\label{fig:tSGM_refined_tSGM_trained}}
    \vspace*{5mm}
    \caption{After concordant training, ResDepth seamlessly complements any stereo matching method.
    (a)~Ground truth DEM.
    (b)~Initial DEM by \textit{s2p}.
    (c)~Initial DEM by tSGM.
    (d)~\textit{s2p} DEM after refinement with ResDepth-\textit{stereo} trained for tSGM.
    (e)~\textit{s2p} DEM after refinement with ResDepth-\textit{stereo} trained for \textit{s2p}.
    (f)~tSGM DEM after refinement with ResDepth-\textit{stereo} trained for tSGM.}
    \label{fig:tSGM_vs_s2p}
\end{figure*}

Fig.~\ref{fig:tSGM_vs_s2p} depicts a selected area of the test stripe in \roiOne. The initial reconstruction obtained by \textit{s2p} is shown in Fig.~\ref{fig:s2p_initial}. Compared to the tSGM DEM in Fig.~\ref{fig:tSGM_initial}, the initial \textit{s2p} DEM exhibits larger scatter around the correct surface.

\mypara{DEM Refinement}
We test two versions of the ResDepth-\textit{stereo} network. In the first run, we use our pre-trained network and apply it without further fine-tuning to the \textit{s2p} DEM, see Fig.~\ref{fig:s2p_refined_tSGM_trained}. As the model has never seen any \textit{s2p} DEM sample during training, it is not able to fully compensate for the error and biases of the \textit{s2p} matcher. After retraining on the training stripes of \roiOne and the initial \textit{s2p} DEM, the model better adapts to the specific error patterns of the initial DEM, see Fig.~\ref{fig:s2p_refined_s2p_trained}. The reconstruction quality is within the expected range and comparable to a tSGM-based reconstruction that has been refined with its matched ResDepth-\textit{stereo} network (see Fig~\ref{fig:tSGM_refined_tSGM_trained}).

{\small
\bibliographystyleSupp{ieee_fullname}
\bibliographySupp{egbib}
}

\end{document}